\documentclass[10pt,journal]{IEEEtran}

\usepackage{balance}
\usepackage{textcomp}
\usepackage{stfloats}
\usepackage{url}
\usepackage{verbatim}
\usepackage{microtype}
\usepackage{graphicx}
\usepackage{subfigure}
\usepackage{booktabs} 
\usepackage{amsmath,amsfonts,amssymb,amsbsy,bm,paralist,theorem,ifthen,color}
\usepackage{algorithmic}
\usepackage{algorithm}
\usepackage{hyperref}
\usepackage{multirow}
\usepackage{cuted}
\usepackage{booktabs}
\usepackage{threeparttable}

\newcommand\Dc{\ensuremath{\mathcal{D}}}

\newcommand\Sc{\ensuremath{\mathcal{S}}}

\newcommand\Lc{\ensuremath{{\mathcal{L}}}}

\newcommand\Bc{\ensuremath{{\mathcal{B}}}}

\newcommand\xb{\ensuremath{{\bf x}}}

\newcommand\yb{\ensuremath{{\bf y}}}

\newcommand\wbb{\ensuremath{\boldsymbol{w}}}

\newcommand\nub{\ensuremath{{\bm \nu}}}
\newcommand\lambdab{\ensuremath{{\bm \lambda}}}

\newcommand\xib{\ensuremath{{\bm \xi}}}

\newcommand\E{\ensuremath{{\mathbb{E}}}}

\newcommand\Xc{\ensuremath{{\mathcal{X}}}}
\newcommand{\wt}{\widetilde}

\newtheorem{Lemma}{Lemma}

\newtheorem{Theorem}{Theorem}
\newtheorem{Def}{Definition}

\newtheorem{Remark}{Remark}
\newtheorem{assumption}{Assumption}

\ifCLASSOPTIONcompsoc
 \usepackage[nocompress]{cite}
\else
  \usepackage{cite}
\fi

\ifCLASSINFOpdf
\else
\fi
\hypersetup{colorlinks=true,linkcolor=black,citecolor=black,urlcolor=black}
\def\BibTeX{{\rm B\kern-.05em{\sc i\kern-.025em b}\kern-.08em
    T\kern-.1667em\lower.7ex\hbox{E}\kern-.125emX}}

\allowdisplaybreaks[4]
\begin{document}
%
\title{Federated Stochastic Primal-dual Learning with Differential Privacy}
%
%
%
%

\author{Yiwei~Li,~\IEEEmembership{Student Member,~IEEE,}
         ~Shuai~Wang,~\IEEEmembership{Member,~IEEE,} \\
        ~Tsung-Hui~Chang,~\IEEEmembership{Senior Member,~IEEE,}
        ~Chong-Yung~Chi,~\IEEEmembership{Fellow,~IEEE}
\thanks{The work of C.-Y. Chi is supported by the Ministry of Science and Technology
under   Grant MOST 109-2221-E-007-088 and  MOST 110-2221-E-007-031.   The work of T.-H. Chang is supported in part by the NSFC, China, under
Grant  61571385 and 61731018.}
\IEEEcompsocitemizethanks{\IEEEcompsocthanksitem Y. ~Li and C.-Y.~Chi are with Institute of Communications Engineering, National Tsing Hua University, Hsinchu, Taiwan (e-mail: lywei0306@foxmail.com,~cychi@ee.nthu.edu.tw).\protect
\IEEEcompsocthanksitem S. ~Wang and T.-H.~Chang are with the School of Science and Engineering, The Chinese University of Hong Kong, Shenzhen, and Shenzhen Research Institute of Big Data, Shenzhen, 518172, China (e-mail: shuaiwang@link.cuhk.edu.cn, tsunghui.chang@ieee.org).   }}

\maketitle
\begin{abstract}
Federated learning (FL) is a  new paradigm that enables many clients to jointly train a machine learning (ML) model under the orchestration of a  parameter server while keeping the local data not being exposed to any third party.  However, the training of FL is an interactive process between local clients and the parameter server.  Such process would cause privacy leakage since adversaries may retrieve sensitive information by analyzing the overheard messages. In this paper, we propose a new federated stochastic primal-dual algorithm with differential privacy (FedSPD-DP). Compared to the existing methods, the proposed FedSPD-DP incorporates local stochastic gradient descent (local SGD) and partial client participation (PCP) for addressing the issues of communication efficiency and straggler effects due to randomly accessed clients. Our analysis shows that the data sampling strategy and PCP can enhance the data privacy whereas the  larger number of local SGD steps could increase privacy leakage, revealing a non-trivial tradeoff between algorithm communication efficiency and privacy protection.  Specifically, we show that, by guaranteeing $(\epsilon, \delta)$-DP for each client per communication round,
the proposed algorithm guarantees $(\mathcal{O}(q\epsilon \sqrt{p T}), \delta)$-DP after $T$ communication rounds while maintaining an $\mathcal{O}(1/\sqrt{pTQ})$ convergence rate for a convex and non-smooth learning problem, where $Q$ is the number of local SGD steps, $p$  is the client sampling probability,  $q=\max_{i} q_i/\sqrt{1-q_i}$ and $q_i$ is the  data sampling probability of each client under PCP. Experiment results are presented to evaluate the practical performance of the proposed algorithm and comparison with state-of-the-art methods.
\end{abstract}



\IEEEdisplaynontitleabstractindextext

%
\IEEEpeerreviewmaketitle

\ifCLASSOPTIONcompsoc
\IEEEraisesectionheading{\section{Introduction}\label{sec:introduction}}
\else
\section{Introduction} \label{sec:introduction}
\fi

%
%
%
%
\IEEEPARstart{R}{ecently}, the distributed machine learning (ML) techniques are rapidly developed due to the high demand for large-scale and distributed data processing\cite{chang2020distributed,wang2021quantized, wang2020federated}.   Simultaneously, privacy is becoming a growing concern for distributed ML systems because data are usually collected from private individuals who naturally wish to protect their own privacy. However, much of the attention on conventional ML models have been focused on developing and improving data processing technologies based on the assumption that all clients or servers are trustworthy. However, this assumption proves to be unrealistic in a variety of applications.  For example,  some sensitive data used in many ML models are publicly released or  to a limited number of third parties, potentially exposing some of the private information contained in the original dataset.  Additionally, strong adversaries may exist who can infer or recover the original data using cutting-edge deep learning techniques \cite{geiping2020inverting}.  Also,  some of the honest-but-curious servers may unintentionally infer the data owners' sensitive information. As a result, privacy-preserving in distributed ML is paramount and has recently received significant attention.  FL is an emerging distributed ML paradigm in which many clients collaboratively train a model under the orchestration of a parameter server (PS) while   mitigating privacy risks and costs by sharing only the gradients or model information and leaving the raw training data on devices  \cite{McMahan2016FederatedLO}.

Nevertheless, FL also poses several key challenges that distinguish it from the conventional distributed ML, such as expensive communication costs and the straggler effect \cite{li2019convergence}.  To be specific, $(i)$ communication costs: training an FL algorithm in a distributed manner requires that the participating clients frequently communicate with the PS about their trained parameters via wireless links, which would demand a large amount of wireless resources, especially when the ML model size is large; $(ii)$ straggler effect: unlike traditional distributed ML, FL has no control over the clients, and some may randomly join or leave the training task, resulting in time-varying active clients during the training process. Waiting for all clients' responses in FL is commonly impractical because inactive clients or stragglers would significantly slow down the entire training process. As a result, it is more practical to model the FL training process with a random subset of clients in each communication round; $(iii)$ privacy protection: although the FL has the inherent advantages of not directly sharing the local clients' data. However, the individual sensitive information can still be exposed to adversaries' model inversion attacks \cite{fredrikson2015model,geiping2020inverting}, or differential attacks \cite{dwork2014algorithmic,bagdasaryan2020backdoor}  during the training.   In such attacks, a client's privacy and data information can be revealed by analyzing the differences between the related model parameters uploaded by the clients \cite{bagdasaryan2020backdoor,wang2019beyond,ma2020safeguarding}.

The most well-known algorithm to solve the above FL challenges is federated averaging (FedAvg) \cite{mcmahan2016communication}. FedAvg allows a subset of clients to perform multiple steps of stochastic gradient descent (SGD) (i.e., local SGD) in parallel during each communication round. The global model is then averaged and updated using the local uploaded models by the PS. FedAvg and its variants have the potential to improve communication efficiency and mitigate the straggler effect in FL.   Another class of methods is based on primal-dual optimization.   The work  \cite{zhang2021fedpd} has demonstrated that the federated primal-dual method could solve the FL problems in a more communication efficient manner than FedAvg and other related algorithms \cite{sahu2018convergence}. Currently, a vast work of primal-dual methods have been proposed to solve various distributed ML problems with excellent convergence performance \cite{Shi2014OnTL, Makhdoumi2017ConvergenceRO,Hong2017StochasticPG} but few of them are adapted to the challenges of the FL system.

As mentioned previously,  privacy preservation is a critical issue of the FL systems.  A common approach to preventing differential attacks  is known as differential privacy (DP),  which  can guarantee with high confidence that an adversary cannot deduce any private information from databases or released models \cite{dwork2006our}.  The adversaries could be outsiders who eavesdrop the shared messages or the honest-but-curious server who may be interested in inferring sensitive information about clients.  By using the DP, even if adversaries collect all the intermediate results,  the sensitive information of the users can not be inferred \cite{dwork2014algorithmic}. Although many techniques have been used in FL to provide the privacy guarantee, such as secure multiparty computation (SMC) \cite{mohassel2017secureml}, homomorphic encryption \cite{giacomelli2018privacy}, secret sharing \cite{bonawitz2017practical},   these methods require complicated encryption protocols and computation, imposing prohibitive overheads on local clients and the PS. Particularly when the model size increases, cryptographic-based privacy-preserving techniques become much less feasible for large-scale distributed ML applications since the computation power of local clients is typically limited \cite{vepakomma2018no}.  In contrast, the DP techniques provide the privacy guarantee without additional overheads and have been considered in many FL applications \cite{kairouz2021advances,triastcyn2019federated,truex2020ldp}.

\subsection{Related works}
The DP techniques applied to  FL has been  investigated extensively, which can be summarized  as  follow: differentially private SGD (DP-SGD) \cite{abadi2016deep}, differentially private federated averaging (DP-FedAvg) \cite{li2020secure, wei2019performance} and differentially private  alternating direction method of multipliers   (DP-ADMM)  \cite{huang2019dp}.   

The work of DP-SGD \cite{abadi2016deep} focuses on tighter estimates of the total privacy loss for general deep learning problems but not for FL scenarios.  The work \cite{li2020secure} demonstrated  that DP-FedAvg maintains $\mathcal{O}(1/T)$ convergence rate on strongly convex and smooth FL problems, where $T$ is the communication round.  The privacy-preserving ADMM-based algorithms in FL have been explored recently because of their good convergence performance.  The work \cite{Zhang2017DynamicDP} proposed two noise perturbation methods: primal perturbation and dual perturbation to guarantee DP  in ADMM-based distributed learning.  The work \cite{zhang2018improving}  proposed modified ADMM with DP by introducing a time-varying  penalty parameter in the distributed network, which shows a better tradeoff  between the utility of the algorithm and  privacy preservation over the entire training process.  The work \cite{zhang2018recycled} proposed a recycled ADMM with persevering DP  in the distributed network where the variables at an odd number of iterations could be reused by the even number of iterations. Consequently, half of the updates do not need adding noise perturbation, maintaining the total privacy loss while improving the learning performance at the same time.  The work \cite{guo2018practical} developed an asynchronous ADMM with DP that the users train local models on their own  data and only share part of the local model with the PS. In \cite{huang2019dp}, they  proved that differentially private ADMM  can achieve the convergence rate of $\mathcal{O}(1/\sqrt{T})$ under convex but non-smooth distributed ML problems.  The work \cite{ding2019stochastic} proposed a stochastic DP-ADMM using SGD to reduce the computation complexity in each local iterates. The work \cite{wang2020privacy}  developed an approach that the privacy-preserving is divided into two phases, in which the Gaussian noise is added on both local raw data  and local uploaded model. This approach strengthens privacy protection for sensitive local data whereas degrading the learning performance.

As shown in Table \ref{tab:table1}, most of the aforementioned works are devoted to studying privacy-preserving ADMM algorithms for conventional distributed ML scenarios. In particular, they did not consider the communication efficiency, straggler effect and privacy protection in one work. The limitations of these works motivate us to explore advanced privacy-preserving primal-dual-based algorithms for FL.
\vspace{-0.2cm}
\begin{table}[!t]
\begin{threeparttable}
\centering
\caption{Summary of privacy perserving  algorithms}
\label{tab:table1}
\begin{tabular}{|c|c|c|c|c|}
\hline
Algorithm &
  \begin{tabular}[c]{@{}c@{}}Local\\  $\text{updates}^{1}$\end{tabular} &
    \begin{tabular}[c]{@{}c@{}}Objective \\ function$^{2}$  \end{tabular} &
  \begin{tabular}[c]{@{}c@{}} ${\rm PCP}^{3}$ \end{tabular} &
  \begin{tabular}[c]{@{}c@{}}Convergence\\  rate\end{tabular}
 \\ \hline
DP-SGD \cite{abadi2016deep}        & LSGD  & NC \& S  & $\times$   & -    \\ \hline
DP-FedAvg \cite{li2020secure}        & LMSGD    &  SC \& S & \checkmark   & $\mathcal{O}(\frac{1}{TQ})$    \\ \hline
DP-ADMM \cite{huang2019dp}        & LGD   &  C \& NS & $\times$   & $\mathcal{O}(\frac{1}{\sqrt{T}})$     \\ \hline
DDP-ADMM \cite{Zhang2017DynamicDP} & LGD   & SC \& S   & $\times$   &             -                     \\ \hline
DP-MADMM \cite{zhang2018improving} & LGD    & C \& S &  $\times$   &    $\mathcal{O}(\frac{1}{\sqrt{T}})$      \\ \hline
DP-RADMM \cite{zhang2018recycled}  & LGD  & C \& S &  $\times$  &             -                     \\ \hline
DP-AADMM\cite{guo2018practical}   & LGD    &  C \& NS & \checkmark &             -                        \\ \hline
DP-LR-ADMM\cite{wang2020privacy}    & LGD    & C \& S & $\times$   &             -                         \\ \hline
Our work                & LMSGD  & C \& NS & \checkmark & $\mathcal{O}(\frac{1}{\sqrt{TQ}})$  \\ \hline
\end{tabular}
\begin{tablenotes}
	\item $^{1}$Local updates: LMSGD-Local multiple SGD steps; LGD-Local one GD step; LSGD-Local one SGD step.
    \item $^{2}$Shorthand notation for objective functions: SC: Strongly Convex, C: Convex, NC: Non-Convex,  S: Smooth and NS: Non-Smooth.
    \item $^{3}$PCP: partial clients participation.
    \item DDP-ADMM: Dynamic DP-ADMM.
    \item DP-MADMM: Modified ADMM with DP.
    \item DP-RADMM: Recycled ADMM with   DP.
    \item DP-AADMM: Asynchronous ADMM with   DP.
    \item DP-LR-ADMM: Privacy preserving via local data randomization and ADMM perturbation.
     \end{tablenotes}
\end{threeparttable}
\end{table}

\subsection{Contributions}
In this paper, we propose a new privacy-preserving federated primal-dual algorithm  for solving the convex and non-smooth problem in the FL network. In particular,  we focus on the effect of local SGD,   PCP  and the total privacy loss on the convergence performance. Specifically, under a total privacy loss constraint, both data sampling strategy and PCP significantly effect the convergence speed and learning performance. The main contributions of this paper are summarized as follows:
\begin{itemize}
\item We propose a novel federated stochastic primal-dual with differential privacy (FedSPD-DP) algorithm  that simultaneously considers the challenging issues of communication efficiency, straggler effect and privacy protection in the FL system.

\item We analytically show that the proposed FedSPD-DP algorithm guarantees   $(\mathcal{O}(q\epsilon \sqrt{p T}), \delta)$-DP (cf. Remark \ref{remark:remark_dp}) and maintains an $\mathcal{O}(1/\sqrt{pTQ})$ convergence rate (cf. \eqref{eqn:2022-4-18-1031}) for a convex and non-smooth learning problem.  Moreover,  we  reveal  the non-trivial tradeoff between the privacy protection and algorithm communication efficiency, and also the tradeoff between the learning performance and the total privacy loss.

\item  We conduct extensive simulations to validate the effectiveness of the proposed FedSPD-DP based on real-world datasets.  The experimental results demonstrate that the performance of FedSPD-DP  outperforms most  state-of-the-art algorithms.

\end{itemize}

{\bf Synopsis:}  Section \ref{sec:preliminaries} introduces the preliminaries of FL and DP.   Section \ref{sec:Federated_Learning_with_Stochastic_ADMM} presents the proposed algorithm FedSPD-DP.  Section \ref{sec:Convergence Analysis} presents the convergence analysis of the proposed algorithm.  The experiment results are presented in Section \ref{sec: sim results}, and Section \ref{sec: conclusion} concludes the paper.

{\bf Notation:} We use $\|\cdot\|$ and $\|\cdot\|_1$ to denote the Euclidean norm and $\ell_{1}$-norm for vectors, respectively. The operator $\mathbb{E}[\cdot]$ represents the statistical expectation;  $[N]$ denotes the integer set $\{1,\ldots, N\}$;    $\nabla f_{i}( \xb )$   denotes the full gradient of function $f_{i}( \xb )$;  $\text{dom}(R)$ represents the domain of function $R$;  $\langle \xb,   \yb \rangle = \xb^{\top} \yb $ represents the inner product operator, where the superscript $`\top$'  denotes the  vector transpose;  $ \mathbb{Z}^{+}$ denotes the positive integer set.   $\operatorname{prox}_{\alpha R} (\boldsymbol{v})$ is denoted as the proximal operator for a convex (possibly  non-smooth) function $R$ \cite{parikh2014proximal}, i.e.,
$$
\operatorname{prox}_{\alpha R} (\boldsymbol{v})=\arg \min\limits_{\xb} \frac{1}{2\alpha}\|\xb- \boldsymbol{v}\|^{2} + R (\xb),
$$
where $\alpha > 0 $ is a constant.

%

\section{Problem Statement and Preliminaries} \label{sec:preliminaries}
We introduce the preliminaries about FL in Sec. \ref{subsec:Federated Learning},  and present the overview of the primal-dual method in  Sec. \ref{subsec:Federated_Learning_with_ADMM}. The background of DP is introduced in   Sec. \ref{subsec:Differential Privacy}.
\vspace{-0.3cm}
\subsection{Federated Learning (FL)} \label{subsec:Federated Learning}
\begin{figure}[t]
\begin{center}
\resizebox{0.95\linewidth}{!}{\hspace{-0cm}\includegraphics{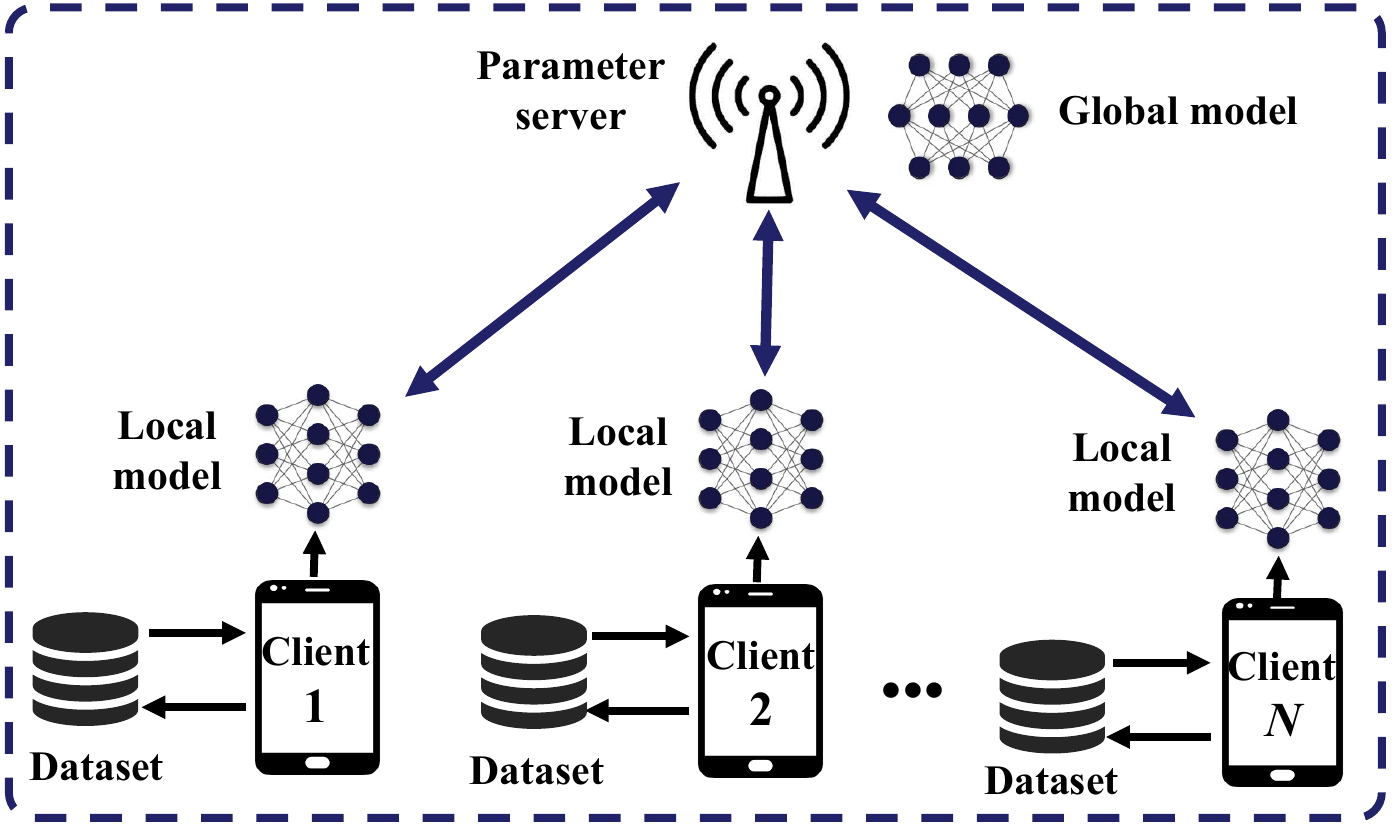}}
\end{center}
\vspace{-0.3cm}
\caption{The framework of FL system.}
\label{fig:fedavg_1}
\end{figure}

Consider an  FL network consisting of a parameter server (PS) and a set of clients  $[N]$ as shown in Fig. \ref{fig:fedavg_1}. The PS aims to learn an ML model through collaboration with the clients without directly accessing their private raw data.
Suppose that each client $i \in [N]$ holds a private dataset $\mathcal{D}_{i}\triangleq \left\{\big(\boldsymbol{a}_{i,j}, \boldsymbol{b}_{i,j}\big) \right\}_{j=1}^{m_i}$, where $m_i$ is the number of local training samples in  $\mathcal{D}_{i}$,   $\boldsymbol{a}_{i,j} \in \mathbb{R}^{d_1}$ is the
data feature vector of the   $j$-th training data sample, and $\boldsymbol{b}_{i,j} \in \mathbb{R}^{d_2}$ is the corresponding  data label.    The objective of FL is to build an ML model over the aggregated sensitive dataset from data providers through a distributed manner,  which can be formulated as follow.
\begin{align}
\!\!\!(\mathcal{P}1) \ \min_{\xb}   \sum_{i=1}^{N}    F_{i}(\xb) \triangleq  \sum_{i=1}^{N} \big( f_{i}(\xb) +  R_{i}(\xb) \big) \label{eqn:fedavg}
\end{align}
where    $\xb \in \mathbb{R}^{d}$ is the trained ML   model, $f_{i}: \mathbb{R}^{d}   \rightarrow \mathbb{R} $ is a user-specified  convex objective function,   $ R_{i}(\cdot): \text{dom}(R)\rightarrow \mathbb{R} \cup \{\infty\}$ refers to a convex but possibly non-smooth regularizer such as the $\ell_{1}$-norm for sparse models. The local objective function $f_{i}(\xb)$ is defined by
\begin{equation}
\begin{aligned}
f_{i}(\xb) \triangleq \frac{1}{m_{i}}\sum_{j=1}^{m_{i}}  f_{i}(\xb; (\boldsymbol{a}_{i,j}, \boldsymbol{b}_{i,j})).
\end{aligned}
\end{equation}
\vspace{-0.5cm}
\subsection{Preliminaries of Primal-dual Method} \label{subsec:Federated_Learning_with_ADMM}
To apply distributed primal-dual method to the  FL network, we reformulate problem $(\mathcal{P}1)$ as
\begin{subequations} \label{eqn:p3}
\begin{align}
\!\!\!(\mathcal{P}2) \ \min _{\left\{\xb, \xb_{0} \right\}} &   \sum_{i=1}^{N}  \big( f_{i}(\xb_{i}) +    R_{i}(\xb_{i}) \big)  \label{eqn:p3_a} \\
\text { s.t. } &   \xb_{i}=\xb_{0}, \quad \forall i \in [N] \label{eqn:p3_b}
\end{align}
\end{subequations}
where   $\xb_{i} \in \mathbb{R}^{d}$  is the local model, and   $\xb_{0} \in \mathbb{R}^{d}$  is the global model.  The constraints (\ref{eqn:p3_b}) enforce that all the local models reach consensus. 

To solve problem $(\mathcal{P}2)$, we  define the augmented Lagrangian as follows
\begin{small}
\begin{align}
\mathcal{L} \big( \xb, \xb_{0}, \lambdab\big) &=\sum_{i=1}^{N}  \mathcal{L}_{ i}\big(\xb_{i}, \xb_{0}, \lambdab_{i} \big), \label{eqn: augmented Lagrangian_1}
\end{align}
\end{small}
where
\begin{small}
\begin{align}
\mathcal{L}_{i}\big(\xb_{i}, \xb_{0}, \lambdab_{i}\big) = & f_{i}\big(\xb_{i}\big) +    R_{i}\big(\xb_{i}\big) +  \langle \lambdab_{i}, \xb_{0} -  \xb_{i} \rangle  \notag\\
&~~+  \frac{\rho}{2}  \left\|\xb_{i} - \xb_{0}\right\|^{2}. \label{eqn: augmented Lagrangian_2}
\end{align}
\end{small}In \eqref{eqn: augmented Lagrangian_2}, $\rho >0$ is the penalty parameter, $\lambdab=[\lambdab_{1}^{\top},\ldots, \lambdab_{N}^{\top}]^{\top} \in \mathbb{R}^{d \times N}$ are the dual variables associated with the constraint \eqref{eqn:p3_b}. The standard distributed primal-dual method, i.e., ADMM \cite{Hong2017StochasticPG}, solves (\ref{eqn:p3})   by minimizing (\ref{eqn: augmented Lagrangian_2}) with respect to (w.r.t.)  $\xb$ and   $\xb_0$ alternatively, followed by the dual variable update of $\lambdab$. Specifically, for iteration $t=0, 1, \ldots $, the updating steps are listed as follows:
\begin{subequations}\label{eqn:admm-update1}
\begin{align}
\xb_{0}^{t} &= \arg \min\limits_{ \xb_{0}} \mathcal{L} \big(\xb^{t-1}, \xb_{0}, \mathbf{\lambdab}^{t-1}\big),  \label{eqn: admm-update_a}\\
\xb_{i}^{t} &= \arg \min\limits_{ \xb_{i}} \mathcal{L}_{i}\big(\xb_{i}, \xb_{0}^{t}, \lambdab_{i}^{t-1}\big),  i \in [N],\label{eqn: admm-update_b}\\
\lambdab_{i}^{t} &= \lambdab_{i}^{t-1} + \rho\big(\xb_{0}^{t} - \xb_{i}^{t}\big), i \in [N]. \label{eqn: admm-update_c}
\end{align}
\end{subequations}
The  above (\ref{eqn: admm-update_b})  and (\ref{eqn: admm-update_c}) are for local model update and the dual variable update, which are performed by distributed  clients, while   (\ref{eqn: admm-update_a}) is the global model update performed by the PS.  Nevertheless, the conventional primal-dual method may not effectively  solve the  FL problem since the update of $\xb_{i}^{t}$ in  (\ref{eqn: admm-update_b}) may not easily be solved due to the complexity of $f_{i}(\cdot)$ and the non-smooth of regularizer term in the objective function.  Like \cite{Hong2017StochasticPG,huang2019dp}, one may consider an inexact update by approximating  the  augmented Lagrangian function, which is given by
\begin{small}
\begin{align}\label{eqn: augmented Lagrangian_1}
 \wt{\mathcal{L}} \big( \xb, \xb_{0}, \lambdab,  \xb^{t-1}\big) &=\sum_{i=1}^{N}  \wt{\mathcal{L}}_{i}\big(\xb_{i}, \xb_{0}, \lambdab_{i}, \xb_{i}^{t-1}\big),
\end{align}
\end{small}
where
\begin{small}
\begin{equation}\label{eqn: augmented Lagrangian_3}
\begin{aligned}
& \wt{\mathcal{L}}_{i}\big(\xb_{i}, \xb_{0}, \lambdab_{i}, \xb_{i}^{t-1}\big) =  f_{i}\big(\xb_{i}^{t-1}\big)  + \langle \lambdab_{i}, \xb_{0}- \xb_{i} \rangle    \\
& ~~+  \langle \nabla f_{i} \big(\xb_{i}^{t-1}; B_{i}^{t}\big), \xb_{i}- \xb_{i}^{t-1} \rangle + \frac{\gamma_i^t}{2}\|\xb_{i} - \xb_{i}^{t-1}\|^{2} \\
& ~~+  \frac{\rho}{2}  \left\|\xb_{i} - \xb_{0}\right\|^{2}+   R_{i}\big(\xb_{i}\big).
\end{aligned}
\end{equation}
\end{small}In \eqref{eqn: augmented Lagrangian_3},   $\gamma_i^t$ is the time-varying parameter (inverse step size), and   $\nabla f_{i} \big(\xb_{i}^{t-1}; B_{i}^{t}\big)=\frac{1}{b}\sum_{j=1}^{b} \nabla f_{i}\big(\xb_{i}^{t-1}; (\boldsymbol{a} _{i,j}^{t}, \boldsymbol{b}_{i,j}^{t})\big)$  is the mini-batch gradient of $ f_i(\xb)$  using a mini-batch data $B_{i}^{t}$ with batch size $b$. By introducing the approximated Lagrangian function \eqref{eqn: augmented Lagrangian_3},  $\xb_{i}^{t}$ can be efficiently updated by proximal gradient descent \cite{parikh2014proximal} since  the approximation enforces the strongly convex of the Lagrangian function.  The updating steps in \eqref{eqn:admm-update1} becomes,
\begin{small}
\begin{subequations} \label{eqn:admm-update2}
\begin{align}
\xb_{0}^{t} &= \arg \min\limits_{ \xb_{0}} \wt{\Lc} \big(\xb_{i}^{t-1}, \xb_{0}, \mathbf{\lambdab}_{i}^{t-1}; \xb_{i}^{t-1}\big)\notag\\
&\overset{(a)}{=}\sum_{i=1}^{N} \Big(\xb_{i}^{t-1} - \frac{\lambdab_{i}^{t-1}}{\rho} \Big) \triangleq \sum_{i=1}^{N} {\bm y}_{i}^{t-1},   \label{eqn: admm-update2_a}\\
\xb_{i}^{t} &= \arg \min\limits_{ \xb_{i}} \wt{\Lc}_{i}\big(\xb_{i}, \xb_{0}^{t}, \lambdab_{i}^{t-1}; \xb_{i}^{t-1}\big),  i \in [N],  \notag\\
&= \text{prox}_{ (\gamma_i^t + \rho )^{-1}   R_{i} }  \Big(  (\gamma_i^t + \rho )^{-1} \big(\gamma_i^t  \xb_{i}^{t-1} + \rho \xb_{0}^{t}  \notag \\
&~~ +   \lambdab_{i}^{t-1} -   \nabla f_{i} \big(\xb_{i}^{t-1};  B_{i}^{t}\big)\big)\Big), i \in [N], \label{eqn: admm-update2_b} \\
\lambdab_{i}^{t} &= \lambdab_{i}^{t-1} + \rho\big( \xb_{0}^{t} - \xb_{i}^{t} \big), i \in [N], \label{eqn: admm-update2_c}
\end{align}
\end{subequations}
\end{small}where $(a)$ holds from $\nabla \wt{\Lc}_{\xb_0} \big(\xb_{i}^{t-1}, \xb_{0}^t, \mathbf{\lambdab}_{i}^{t-1}; \xb_{i}^{t-1}\big)=\bf{0}$.

\subsection{Differential Privacy} \label{subsec:Differential Privacy}
In the above primal-dual algorithm, the local model ${\bm y}_{i}^{t}$ will be exchanged with the PS. In this process, the adversaries may reveal the client’s private information by model inversion attacks  from the uploaded models.  To mitigate such attacks and provide the privacy guarantee for local clients, we adopt the  $(\epsilon, \delta)$-DP to protect the local models, which is defined as follows:
\begin{Def}\label{Def:DP defintion}  $(\epsilon, \delta)$-DP {\rm \cite{huang2019dp}}.
Consider two neighboring datasets  $\mathcal{D}$ and $\mathcal{D}^{\prime}$, which  differ in only one data sample. A randomized mechanism $\mathcal{M}$ is  $(\epsilon, \delta)$-DP if for any two $\mathcal{D}$, $\mathcal{D}^{\prime}$ and measurable subset    $S \subseteq Range(\mathcal{M})$, we have
\begin{align} \label{eqn:DP_def}
\operatorname{Pr}[\mathcal{M}(\mathcal{D}) \in S] \leq \exp(\epsilon) \cdot \operatorname{Pr}\left[\mathcal{M}( \mathcal{D}^{\prime}) \in S\right]+\delta,
\end{align}
where $\epsilon > 0$ represents privacy protection level, and $\delta$ stands for the probability to break $\epsilon$-DP.
\end{Def}
A smaller $\epsilon$ means stronger privacy protection.  The intuition of  \eqref{eqn:DP_def} is that it has a negligible effect on the output of  $\mathcal{M}$ when any  individual data sample in the dataset $\mathcal{D}$ is removed or replaced.

\begin{Lemma}  \label{Lemma: global sensitivity}
\cite[Theorem 3.22]{dwork2014algorithmic}   Suppose the query function $g$  access the dataset $\Dc$ via randomized mechanism  $\mathcal{M}$. Then,  the required noise scale adding to $g$ to  guarantee $\big(\epsilon, \delta \big)$-DP is given by
\begin{align}
\sigma = \frac{2 s^{2} \ln(1.25/\delta)}{\epsilon^{2}},
\end{align}where $ s$  is  the  $\ell_2$-sensitivity  of the function $g$ defined by
\begin{equation}\label{eqn:global sensitivity_f}
\begin{aligned}
s \triangleq \max _{\mathcal{D}, \mathcal{D}^{\prime}}\left\|g(\mathcal{D})- g\big(\mathcal{D}^{\prime}\big)\right\|.
\end{aligned}
\end{equation}
\end{Lemma}

\section{Federated stochastic primal-dual algorithm with DP}\label{sec:Federated_Learning_with_Stochastic_ADMM}
This section presents the proposed FedSPD-DP algorithm for FL system and analyze the required noise scale to guarantee $(\epsilon, \delta)$-DP.
\subsection{Proposed FedSPD-DP Algorithm}\label{subsec:Proposed FedSPD-DP}
Based on the primal-dual algorithm update rule in \eqref{eqn:admm-update2},  we design a new scheme to improve the   communication efficiency and solve the straggler effect in the FL system.

\vspace{0.1cm}
{\bf Communication Efficiency:}  Inspired by  \cite{mcmahan2016communication, li2019convergence},   we propose a new scheme that allows the  clients to perform $Q$ steps of local  SGD within one communication round. As a result,  (\ref{eqn: admm-update2_b}) becomes
\begin{small} \label{eqn: local update_primal}
\begin{subequations}
\begin{align}
\xb_{i}^{t,r}  &=  \text{prox}_{ (\gamma_i^t + \rho )^{-1}   R_{i} }  \Big(  (\gamma_i^t + \rho )^{-1} \big(\gamma_i^t  \xb_{i}^{t,r-1} + \rho \xb_{0}^{t}   \notag\\
&~~~~ +   \lambdab_{i}^{t-1} -   \nabla f_{i} \big(\xb_{i}^{t,r-1};  B_{i}^{t,r}\big)\big)\Big),  \label{eqn: local update_primal_1}\\
\xb_{i}^{t} &= \frac{1}{Q}\sum_{r=0}^{Q-1}\xb_{i}^{t,r}, \label{eqn: local update_primal_2}
\end{align}
\end{subequations}
\end{small}where $\xb_{i}^{t,r}$ denotes local model at the $r$-th inner iteration and the $t$-th communication round,    and \eqref{eqn: local update_primal_1} is obtained by replacing ${\bf x}_i^{t-1}$ with ${\bf x}_i^{t,r-1}$ in the first line of \eqref{eqn: admm-update2_a}, and \eqref{eqn: local update_primal_2} is the average of all the obtained models by \eqref{eqn: local update_primal_1}.


To protect the privacy of local model $ {\bm y}_{i}^{t}$    defined in \eqref{eqn: admm-update2_a},  we apply the DP by adding  an artificially generated Gaussian noise    $\boldsymbol{\xi}_{i}^{t} \sim \mathcal{N}\big(\bm{0}, \sigma_{i, t}^{2} \mathbf{I}_{d}\big)$  to the local model, that is,
\begin{small}
\begin{equation}
\begin{aligned} \label{eqn: global_update_partial}
\tilde{{\bm y}}_{i}^{t}  =  {\bm y}_{i}^{t} + \boldsymbol{\xi}_{i}^{t} =    \xb_{i}^{t} - \frac{\lambdab_{i}^{t}}{\rho} + \boldsymbol{\xi}_{i}^{t}, \forall i \in \Sc_t.
\end{aligned}
\end{equation}
\end{small}
{\bf Partial Client Participation (PCP)}
We now elaborate on solving the straggler effect problem in FL system.  Inspired by \cite{mcmahan2016communication},  we propose a new scheme that enables the PS to sample a small subset of clients  at any round $t$   (denoted by $\mathcal{S}_{t}$, with $|\mathcal{S}_{t}|=K $)  and then ask them to upload their local model. Note that,  each client $i \in \mathcal{S}_{t}$ is uniformly sampled without replacement (WOR) with a probability $p_{i} \triangleq  \operatorname{Pr}\big(i \in \mathcal{S}_{t}  \big)$.
The inactive clients (i.e., $i \notin \mathcal{S}_{t}$)  will not update their local model (i.e., $\xb_{i}^{t} = \xb_{i}^{t-1}, \lambdab_{i}^{t} = \lambdab_{i}^{t-1}$) and their latest model will be kept on the PS for the global aggregation together with the perturbed model $\tilde{{\bm y}}_{i}^{t}$ from the selected clients.   It should be emphasized that the model of inactive clients will also be averaged in our scheme, which is quite different from the typical PCP scheme in the FL \cite{mcmahan2016communication}.   The details of the proposed FedSPD-DP  are illustrated in Algorithm \ref{alg:ADMM2}.

We now briefly describe the main steps of  Algorithm \ref{alg:ADMM2}: at  communication round $t$, the PS updates  $\xb_{0}^{t}$  (line 5),  and  randomly samples a subset of clients $\mathcal{S}_{t}$ without replacement   and broadcasts  $ \xb_{0}^{t} $  to all the clients  $i \in  \mathcal{S}_{t}$ (line 6). On the clients side, each of the selected client performs $Q$ steps of local SGD (line 13), and  then $\xb_{i}^{t} $ is obtained by averaging $Q$ steps of local update results (line 15).  After that, the dual variable $\lambdab_{i}^{t} $ is updated (line 16). Based on the result of $\xb_{i}^{t}$ and  $\lambdab_{i}^{t} $, the local  model  $ \tilde{{\bm y}}_{i}^{t}$ is obtained by adding a zero-mean Gaussian noise to $ {\bm y}_{i}^{t}$ (line 18).  For the inactive clients $i \notin  \mathcal{S}_{t}$, their  model will not be updated (line 22).  Finally,  all $ \tilde{{\bm y}}_{i}^{t}, \forall i \in \mathcal{S}_{t}$ are sent to the PS for the next round of aggregation.

Compared with the existing primal-dual methods, our algorithm allows  a  subset of clients to run the $Q$ steps of local proximal SGD  within one communication round.  More importantly, our approach guarantees the client-level   DP while preserving model accuracy and saving communication costs simultaneously.

\begin{algorithm}[!t]
\caption{Federated stochastic primal-dual with DP}
\begin{algorithmic}[1]\label{alg:ADMM2}
\STATE \textbf{Input:}    $b$, $Q$, $K$ and $T$.
\STATE Initialize $\xb_{0}^{0}$, $\{\xb_{i}^{0,0}\}_{i \in [N]}$ and  $\{\lambdab_{i}^{0}\}_{i \in [N]}$.

\FOR{$t=1, \ldots, T$}
\STATE  $\textbf{The PS side:}$
\STATE    Update   $ \xb_{0}^{t} =  \frac{1}{N} \sum_{i=1}^{N} \tilde{{\bm y}}_{i}^{t} $.
\STATE  Update the subset of clients ${\cal S}_t \subseteq [N]$ through uniformly sampling (i.e., PCP), and broadcast ${\bf x}_0^t$ to all clients.
\STATE $\textbf{Client side:}$
 \FOR{$i \in \mathcal{S}_{t}$ in parallel}
 \STATE Initialize $\xb_{i}^{t,0}= \xb_{i}^{t-1,Q}$
   \FOR{$r= 0, \ldots, Q-1$}
    \STATE  Sample mini-batch $B_{i}^{t,r}$  from $\mathcal{D}_{i}$.
   \STATE  Calculate the  stochastic gradient   $\nabla f_{i} \big(\xb_{i}^{t,r-1}; B_{i}^{t,r}\big)$.
   \vspace{-0.35cm}
  \STATE  Update  the  $\xb_{i}^{t,r}$  using \eqref{eqn: local update_primal_1}.
\ENDFOR
    \STATE Compute $\xb_{i}^{t} $ by \eqref{eqn: local update_primal_2}.
     \STATE  Compute $\lambdab_{i}^{t}  = \lambdab_{i}^{t-1}- \rho\big(\xb_{i}^{t}-\xb_{0}^{t}\big)$.
     \STATE Sample $\boldsymbol{\xi}_{i}^{t} \sim \mathcal{N}\big(\bm{0}, \sigma_{i, t}^{2} \mathbf{I}_{d}\big)$.
     \STATE  Compute  $\tilde{{\bm y}}_{i}^{t} $ via \eqref{eqn: global_update_partial}.
    \STATE Send  $ \tilde{{\bm y}}_{i}^{t}$ to the PS for next round of aggregation.
\ENDFOR
 \FOR{$i \notin \mathcal{S}_{t}$ in parallel}
    \STATE Update $ \xb_{i}^{t} =  \xb_{i}^{t-1}$, $\lambdab_{i}^{t}  = \lambdab_{i}^{t-1} $ and $ \tilde{{\bm y}}_{i}^{t} =  \tilde{{\bm y}}_{i}^{t-1}$.
 \ENDFOR
\ENDFOR
\end{algorithmic}
\end{algorithm}

\subsection{Privacy Analysis}\label{subsec:Privacy Analysis}
To further analyze the required noise for providing $(\epsilon, \delta)$-DP at each communication round, we need the following assumptions.
\begin{assumption}\label{Ass: Assumption1}
Each local function $f_i$ is convex and differentiable, and the regularizer $R_i$ is convex but possibly non-smooth.
\end{assumption}

\vspace{-0.3cm}
\begin{assumption}\label{Ass: Assumption2}  Denote $(\boldsymbol{a}_{i}^{t}, \boldsymbol{b}_{i}^{t})$ as one data sample and  $\nabla f_{i}\big(\xb_{i}^{t}; (\boldsymbol{a}_{i}^{t}, \boldsymbol{b}_{i}^{t})\big)$ as   the   associated   gradient. Then, the following equations are assumed to  hold.
\begin{equation}
\begin{aligned} \label{eqn: p11}
\mathbb{E} \left[\nabla f_{i}\big(\xb_{i}^{t}; (\boldsymbol{a}_{i}^{t}, \boldsymbol{b}_{i}^{t}) \big) \right] &=  \nabla f_{i}\big(\xb_{i}^{t} \big),  \\
\mathbb{E} \big[ \big\|\nabla f_{i}(\xb_{i}^{t}; (\boldsymbol{a}_{i}^{t}, \boldsymbol{b}_{i}^{t}) )-\nabla f_{i}(\xb_{i}^{t})\big\|^{2}\big] &\leq \phi^{2}, \forall i\in [N],
\end{aligned}
\end{equation}
\end{assumption}
for some constant $\phi$. Note that the variance of $\nabla f_i({\bf x}_i^t;B_i^t)$ reduces to $\phi^{2}/b$ for mini-batch SGD with batch size $b$.
\begin{assumption}\label{Ass: Assumption3}
(Bounded Domain)   Assume that $\rm{dom} (R)$ and $\mathcal{B}_{\bf \lambda} \triangleq \{\lambdab   \mid \| \lambdab \| \leq \beta \}$ are bounded sets.  Then, there exist finite $d_{\mathcal{X}} >0$  and $d_{\lambda} >0$ satisfying
\begin{align}
d_{\mathcal{X}} &= \sup_{\xb, \hat{\xb} \in \rm{dom} (R)} \big\| \xb - \hat{\xb}  \big\|, \label{eqn:assumption_bounded_domain}\\
d_{\lambda} &= \sup_{ \lambdab, \hat{\lambdab}  \in \mathcal{B}_{\bf \lambda}} \big\| \lambdab -  \hat{\lambdab} \big\|. \label{eqn:bounded_dual}
\end{align}
\end{assumption}
Based on Assumptions \ref{Ass: Assumption1}-\ref{Ass: Assumption3}, we have the following remark.
\vspace{-0.2cm}
\begin{Remark} \label{remark:remark1}
Suppose that  Assumptions \ref{Ass: Assumption1}-\ref{Ass: Assumption3} hold.  For any $i \in [N]$,  the following inequalities hold
\begin{equation}
\begin{aligned} \label{eqn:bounded_gradient}
\big\|\nabla f_{i}\big(\xb_{i}^{t}; B_{i}^{t}\big) \big\|  &\leq G, \forall i \in [N], \\
\big\| R_{i}\big(\xb_{i}^{t} \big)\big\|   &\leq G_{R}, \forall i \in [N],
\end{aligned}
\end{equation}
for some constants $G$ and $ G_{R}$.
\end{Remark}
To proceed, we next determine the sensitivity of the local  model $ {\bm y}_{i}^{t} $, which is given as in the following lemma.
\begin{Lemma}\label{lemma:sensitivity} Suppose that Assumption \ref{Ass: Assumption3} holds. For any $t$ and $i \in [N]$,  the sensitivity  for  ${\bm y}_{i}^{t} $  (cf. \eqref{eqn:global sensitivity_f}) is given by
\begin{align}
s_{i,t}   = \left\{\begin{array}{ll}   \frac{ 4G}{ (\rho +\gamma_i^t)},     & {\text { when} \ Q=1,}    \\
 \frac{ 4QG}{ (Q-1) (\rho +\gamma_i^t)},     & {\text { when} \ Q>1.}   \end{array}\right.
\end{align}
\end{Lemma}
\textit{Proof:} See the Appendix \ref{appdix: proof of Lemma}.

Then, according to Lemma \ref{Lemma: global sensitivity} and Lemma \ref{lemma:sensitivity}, the required noise scale $\sigma_{i,t} $ for guaranteeing $(\epsilon, \delta )$-DP    is
\begin{align}\label{eqn:noise for with replacement_0}
\sigma_{i,t}  = \left\{\begin{array}{ll}  \frac{  4G \sqrt{2\ln \big(1.25/\delta\big)}}{  \epsilon \big( \rho +\gamma_i^t\big)},     & {\text { when} \ Q=1,}    \\
\frac{  4QG \sqrt{2\ln \big(1.25/\delta\big)}}{ (Q-1) \epsilon \big( \rho +\gamma_i^t\big)},     & {\text { when} \ Q>1.}   \end{array}\right.
\end{align}
The result of (\ref{eqn:noise for with replacement_0}) shows that the  noise scale $\sigma_{i,t} $   decreases with the communication round  $t$ when  $\gamma_i^t$ increases with $t$.  Recall that the added noise will deteriorate the learning performance.  Hence, it is desirable to have diminishing DP noise with the communication round $t$.
\subsection{Total Privacy Loss Analysis}\label{sec:Total privacy leakage}
In this section,  we calculate the total privacy loss of the proposed FedSPD-DP after $T$ communication rounds using the moments accountant method  \cite{abadi2016deep}.
\begin{Theorem}\label{Thm:total_budget}
Suppose that  any client $i, \forall i \in [N]$ in   Algorithm \ref{alg:ADMM2} is uniformly sampled with   probability $p_i$ and   each round  guarantees $(\epsilon, \delta)$-DP.  The total privacy loss  for  client  $i$ after $T$ communication rounds satisfies
\begin{small}
\begin{align} \label{eqn:total_privacyloss}
\bar{\epsilon}_{i} = c_{0} q_i \epsilon \sqrt{\frac{p_i T}{1-q_i} }, \forall i \in [N],
\end{align}
\end{small}
\vspace{-0.1cm}
where $ c_{0}$ is a constant and $q_i$ is given by
\begin{align}\label{eqn: q_1}
q_i = \left\{\begin{array}{ll}   \frac{Q b}{m_{i}},     & {\text { data sampling WOR,}}    \\
1-(1-\frac{1}{m_{i}})^{Qb},     & {\text { data sampling WR,}}   \end{array}\right.
\end{align}
where WOR and WR denote the data sampling without and with replacement, respectively.
\end{Theorem}
\textit{Proof:} The proof mainly follows the work \cite{huang2019dp}.  However, it did not consider the PCP and local SGD.   To the completeness, we restate the proof in the Appendix \ref{subsec:the_Proof_of_Theorem_dp}.
\begin{Remark} \label{remark:remark_dp}
When the participated clients are uniformly  sampled  by the PS, Theorem \ref{Thm:total_budget} demonstrates that Algorithm \ref{alg:ADMM2}  guarantees $(\mathcal{O}(q\epsilon \sqrt{p T}), \delta)$-DP, where
\begin{align}
q=\max_{i} \frac{q_i}{\sqrt{1-q_i}}, \ p= {p_i}, \forall i \in [N].
\end{align}
Moreover, since $1-(1-\frac{1}{m_{i}})^{Qb} \leq \frac{Q b}{m_{i}}$,   the total privacy loss for the data sampling  WR is smaller than that for the data sampling WOR.
\end{Remark}
%


\section{Convergence Analysis}\label{sec:Convergence Analysis}
In this section, we discuss the convergence performance of the proposed FedSPD-DP.  To proceed, we define the following notations:
\begin{small}
\begin{equation} \notag
\bar{\xb}_{i}^{T} = \frac{1}{TQ}   \sum_{t=1}^{T} \sum_{r=0}^{Q-1} \xb_{i}^{t, r},  \quad \bar{\xb}_{0}^{T}=\frac{1}{T} \sum_{t=1}^{T} \xb_{0}^{t},  \quad \bar{\lambdab}_{i}^{T}=\frac{1}{T} \sum_{t=1}^{T} \lambdab_{i}^{t}.
\end{equation}
\end{small}
A new function  $J(\xb, \xb_{0}, \lambdab)$   is defined as below
\begin{small}
\begin{equation}
\begin{aligned}\label{eqn:J}
J(\xb, \xb_{0}, \lambdab)  &\triangleq \sum_{i=1}^{N} J_{i} (\xb_{i}, \xb_{0}, \lambdab_{i})\\
& = \sum_{i=1}^{N} \frac{1}{p_{i}} \big( F_{i}(\xb_{i})+ \langle -\lambdab_{i}, \xb_{i} - \xb_{0} \rangle \big),
\end{aligned}
\end{equation}
\end{small}where $F_{i}( \xb_{i}) =    f_{i}( \xb _{i}) +   R_{i}( \xb_{i}) $.   To measure the convergence performance, we use the following criterion that has been widely used in previous works \cite{Hong2017StochasticPG},  \cite{huang2019dp},\cite{ouyang2013stochastic}.
\begin{small}
\begin{equation}\label{eqn:criterion}
\!\!\!H(\bar{\xb}_{i}^{T}, \xb_{0}^{T}) \triangleq  \sum_{i=1}^{N} \E \big[F_{i}(\bar{\xb}_{i}^{T})- F_i(\xb_{0}^{\star}) \big]+  \beta  \sum_{i=1}^{N} \E \big[\|\bar{\xb}_{i}^{T}- \xb_{0}^{T}\| \big],
\end{equation}
\end{small}where  $\beta$ is the radius of the ball ${\cal B}_{\bm \lambda}$ in  Assumption \ref{Ass: Assumption3},   $\xb_{0}^{\star}$ is the optimal solution of problem \eqref{eqn:p3}, $ \sum_{i=1}^{N} (F_{i}(\bar{\xb}_{i}^{T}) - F_i(\xb_{0}^{\star}))$ denotes the distance between the achieved objective value and the optimal objective value after $T$ communication rounds.  $ \sum_{i=1}^{N} \| \bar{\xb}_{i}^{T}- \xb_{0}^{T}\|$ measures the difference between the local iterates and the global one. Therefore, when  $H(\bar{\xb}_{i}^{T}, \xb_{0}^{T})=0$,  the algorithm converges to the optimal solution and all local iterates reach consensus.

\subsection{Convergence Result} \label{subsec: convergence performance analysis}
The convergence property for FedSPD-DP is shown in the following theorem.
\vspace{-0.3cm}

\begin{Theorem}\label{Thm:dp theorem2}
Suppose   Assumptions \ref{Ass: Assumption1}-\ref{Ass: Assumption3} hold and the client $i \in [N]$ guarantees the $(\epsilon, \delta)$-DP in each communication round. Moreover, let  $ \rho \geq   \sqrt{\gamma_i^{t}}, \forall t$, and
\begin{small}
\begin{equation}
\begin{aligned}\label{eqn:gamma}
\gamma_i^t & =  \frac{2\sqrt{Q p_i \cdot C(Q,\epsilon)   } }{d_{\mathcal{X}}} \sqrt{t},
\end{aligned}
\end{equation}
\end{small}
where
\begin{small}
\begin{equation}
\begin{aligned}\label{eqn:gamma}
C(Q,\epsilon)     = G^{2}  + 2d_{\lambda}^{2} +   \frac{2\phi^{2}}{b}    +  \frac{16 \rho d G^{2} \ln (1.25/\delta)}{ (Q-1)^2 \epsilon^{2}}.
\end{aligned}
\end{equation}
\end{small}Then, when $Q>1$, we have
\begin{small}
\begin{align}
& \sum_{i=1}^{N}\E \big[ F_{i}(\bar{\xb}_{i}^{T})- F_i(\xb_{0}^{\star})] +  \beta  \sum_{i=1}^{N} \E[\|\bar{\xb}_{i}^{T}- \xb_{0}^{T}\| \big]  \notag\\
\leq &   \frac{1}{T}  \sum_{i=1}^{N}  \sup_{\lambdab_{i} \in \mathcal{B}_{\bf \lambda}}  \E \big[J_{i} (\xb_{i}^{0}, \xb_{0}^{0}, \lambdab_{i}) - J_{i} (\xb_i^{T}, \xb_0^{T}, \lambdab_{i}) \big]  \notag \\
		& +  \frac{1}{2TQ}  \sum_{i=1}^{N} \Big(    \frac{\rho  d_{\mathcal{X}}^{2} (Q+1) }{p_{i}} +     \frac{Q}{p_{i}} \frac{d_{\lambda}^{2}}{\rho}      +  \frac{4 G_{R}}{p_i} \Big) \notag \\
		& +   \sum_{i=1}^{N} \frac{ 2d_{\mathcal{X}} \sqrt{ C(Q,\epsilon)   } }{  \sqrt{ T Q p_i}}. \label{eqn:convergence_result_final}
\end{align}
\end{small}Similarly, when $Q=1$,   \eqref{eqn:convergence_result_final}  is still true   and
\begin{small}
\begin{equation}
\begin{aligned}\label{eqn:gamma}
C(Q,\epsilon)    = G^{2}  + 2d_{\lambda}^{2} +   \frac{2\phi^{2}}{b}   +  \frac{ 16 \rho d G^{2} \ln (1.25/\delta)}{\epsilon^{2} }.
\end{aligned}
\end{equation}
\end{small}
\end{Theorem}
\textit{Proof:} please see  Appendix \ref{appdix: proof of thm 2}.

By Theorem 2, when the active clients are uniformly sampled,  the proposed algorithm (Algorithm \ref{alg:ADMM2}) achieves the order in convergence rate through $T$ communication rounds as follows
\begin{align}\label{eqn:2022-4-18-1031}
	{\cal O}(\frac{1}{\sqrt{pTQ}})~ \text{as $p_i=p$ for all $i\in [N]$,}
\end{align}
thereby showing the impact of client sampling on the learning performance beyond most state-of-the-art algorithms.

\vspace{-0.2cm}
\begin{Remark} \label{remark:remark_impact_of_DP}
({Impact of DP}) Theorem \ref{Thm:dp theorem2}  shows that  there exists a tradeoff between the learning performance and privacy protection level according to \eqref{eqn:convergence_result_final}, that is,  the stronger privacy protection (smaller $\epsilon$) will downgrade  learning performance.  Furthermore, a smaller  $p_i$   will slow down the convergence rate, but  the total privacy loss  will be reduced when    $p_i$  is small as shown in \eqref{eqn:total_privacyloss}.  Thus, there also exists a tradeoff between the convergence performance and the total privacy loss.
\end{Remark}
Furthermore, by considering the convergence rate, the dominating term in  \eqref{eqn:convergence_result_final} is
\begin{small}
\begin{align} \label{eqn:dominating term}
\sum_{i=1}^{N} \frac{ 2d_{\mathcal{X}} \sqrt{ C(Q,\epsilon)   } }{  \sqrt{ T Q p_i}}.
\end{align}
\end{small}Let $T_{\varepsilon}$ denote the number of required communication rounds for the algorithm to  achieve an $\varepsilon$ performance accuracy (i.e., $H(\bar{\xb}_{i}^{T_{\varepsilon}}, \xb_{0}^{T_{\varepsilon}}) \leq \varepsilon$).    By   \eqref{eqn:dominating term},    we have $T_{\varepsilon}$ as follows
\begin{equation}
\begin{aligned} \label{eqn:required communication round}
T_{\varepsilon} \propto \mathcal{O}\big( \frac{1}{Q^{3} \varepsilon^{2} } \big).
\end{aligned}
\end{equation}
By combining \eqref{eqn:total_privacyloss}, \eqref{eqn: q_1}   and \eqref{eqn:required communication round},    and  under data sampling WOR,
the total privacy loss for client $i$ is given by
\begin{equation}
\begin{aligned}\label{eqn:required total_privacy_loss}
\bar{\epsilon}_i  \propto  \mathcal{O}\Big( \frac{Q^{\frac{3}{2}}}{\sqrt{1-\frac{Qb}{m_i}}}  \frac{\sqrt{ p_i}   \epsilon}{m_i \varepsilon}   \Big), \forall i \in [N].
\end{aligned}
\end{equation}
Based on \eqref{eqn:dominating term}, \eqref{eqn:required communication round} and \eqref{eqn:required total_privacy_loss}, we have the following Remarks.

\vspace{-0.1cm}
\begin{Remark} \label{remark:remark_impact_of_localSGD}
({Impact of local SGD}) Under the $\varepsilon$ performance accuracy for the proposed algorithm, larger $Q$ will lead to $(i)$ smaller $T_{\varepsilon}$ (better communication efficiency), thus consistent with \cite{yang2021achieving}; $(ii)$ larger total privacy loss $\bar{\epsilon}_i$, thereby showing the tradeoff between $T_{\varepsilon}$ and $\bar{\epsilon}_i$. However, a larger Q may slow down the convergence speed of the FedAvg algorithm when the local data are highly unbalanced \cite{li2019convergence}. The corresponding study of the proposed algorithm for highly unbalanced data is nontrivial and may be worth future investigation.
\end{Remark}
\begin{Remark} \label{remark:remark_impact_of_PCP}
({Impact of PCP})  PCP will slow down the convergence due to $p_i \leq 1$ (cf. \eqref{eqn:convergence_result_final}).  However,  when the privacy loss  (denoted as $\bar{\epsilon}_{0}$) for the FL system is constrained,  and the learning process will be forced to stop before converging to the optimal solution  when the total privacy loss  $\bar{\epsilon}_i \geq \bar{\epsilon}_{0}, \forall i \in [N]$. For the PCP, inactive clients will not cost any privacy loss, thereby slowing down the expansion of the total privacy loss budget compared to the full client participation (FCP). In other words,  the learning process can still proceed as long as some clients with the nonzero balance of the total privacy loss ($\bar{\epsilon}_i \leq \bar{\epsilon}_{0}$) remain active in the FL system, and thus continuely improving the learning performance.
\end{Remark}

\section{Experiment Results} \label{sec: sim results}
In this section, we evaluate the performance of the proposed   FedSPD-DP  by considering  logistic regression problem with $\ell_{1}$-norm regularizer, i.e., $R(\xb) =  \| \xb \|_{1}$. The local objective is
\begin{small}
\begin{equation}
\begin{aligned} \label{eqn:empirical_loss}
\!\!\!\!\!F(\xb) =  \sum_{i=1}^{N} \sum_{j=1}^{m_{i}} \frac{1}{m_{i}} \big[ \ln \big(1+\exp \big(-\boldsymbol{b}_{i, j} {\bf X}  \boldsymbol{a}_{i, j}\big)\big) \big] +  \lambda_{R} \| \xb \|_{1},
\end{aligned}
\end{equation}
\end{small}where ${\bf X} \in {\mathbb R}^{d_1\times d_2}$ such that ${\bf x} \in{\mathbb R}^{d_1 d_2}$ is the column vector by stacking all the columns of ${\bf X}$,  and $\lambda_{R}$ is the regularizer parameter. Note that, $F({\bf x})$  is a convex but non-smooth function.  The augmented Lagrangian function value (ALFV) defined in \eqref{eqn: augmented Lagrangian_1} and testing accuracy (learning accuracy over the testing data) are evaluated as the learning performance in the simulation, because ALFV, which will monotonically decrease  with the communication round under sufficient large parameter $\rho$,   has been used for measuring the convergence performance in  other primal-dual works \cite{zhang2021fedpd,huang2019dp}.


$\textbf{Datasets: }$ The benchmark dataset Adult \cite{blake1998uci} is used, which consists of 32561 training samples and 16281 testing samples.  In the simulation, the most frequently occurring feature value is used for the missing values in the dataset, and then all the features are normalized.  All the training samples are uniformly distributed among the total of $N = 100$ clients.

$\textbf{Baseline algorithms: }$  We compare the proposed FedSPD-DP with some existing state-of-the-art approaches, including DP-SGD \cite{abadi2016deep}, DP-FedAvg \cite{li2020secure}, and DP-ADMM \cite{huang2019dp}.

$\textbf{Setup: }$  We set the total number of communication rounds  $T=100$ and the penalty parameter $\rho=20$, and set $\delta=10^{-4}$, $\lambda_{R}=0.01$,  $\phi = 1$, $d_{\lambda}=1$, $d_{\mathcal{X}}=1$  and $G=1$. Note that, the data dimension is $d=d_1 \times d_2=162$ after data  preprocessing and the client sampling probability $p_i$ is $ K/N$.  By  Theorem \ref{Thm:dp theorem2},  $\gamma_i^t $ set as follows

\vspace{-0.2cm}
\begin{small}
\begin{equation}
\begin{aligned}
\gamma_i^t = 2 \Big( Q  p_i \big(3 +  \frac{2}{b}   +  \frac{51840 \cdot \ln (1.25/\delta)}{ (Q-1)^2 \epsilon^{2} } \big)\Big)^{\frac{1}{2}}  \cdot  \sqrt{t}.
\end{aligned}
\end{equation}
\end{small}Given the total privacy loss $\bar{\epsilon}$ and total communication round $T$ and the parameter $c_{0}=3.04$, the privacy protection level $\epsilon$ for each communication round can be determined by \eqref{eqn:total_privacyloss} as

\vspace{-0.2cm}
\begin{small}
\begin{equation}
\begin{aligned}
\epsilon =  \frac{\bar{\epsilon} \sqrt{1-q_i}}{c_{0} q_i \sqrt{p_i T}}, \forall i \in [N].
\end{aligned}
\end{equation}
\end{small}In DP-SGD, gradient clipping is performed to determine the global sensitivity. The parameter of gradient clipping bound follows the scheme  in \cite{abadi2016deep} by taking the median of the norms of the unclipped gradients.   For DP-ADMM, the subgradient descent method is used   without using multiple local SGD steps.  For DP-FedAvg, the parameters  setting follows as in \cite{li2020secure}. 

\vspace{-0.3cm}
\subsection{Impact of the total privacy loss $\bar{\epsilon}$ }
\begin{figure}[t!]
\begin{minipage}[b]{0.485\linewidth}
\centering
\includegraphics[scale=0.30]{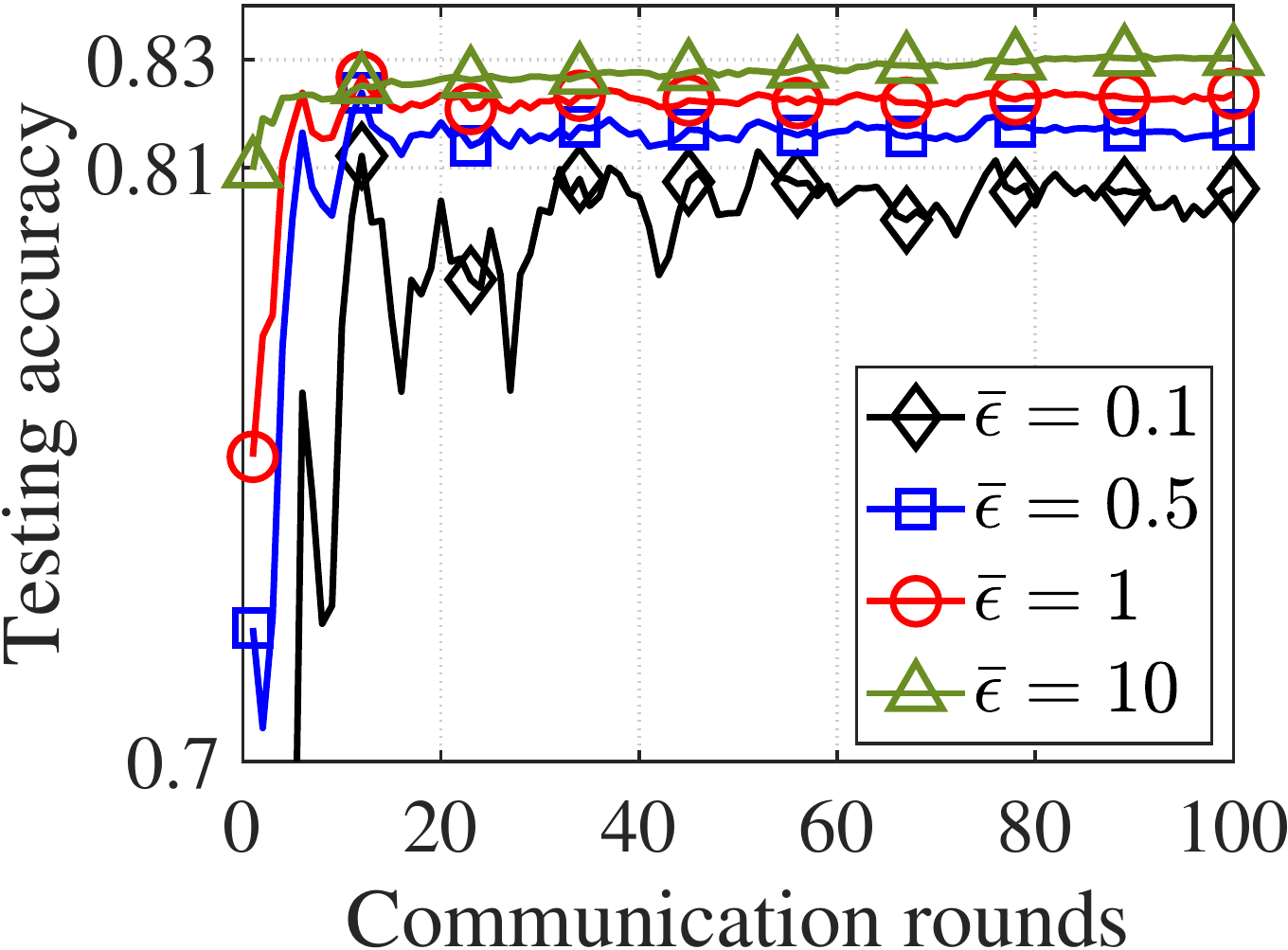} \label{eps_accuracy_a}
\centerline{\scriptsize{(a)}}\medskip
\end{minipage}
\hfill
\begin{minipage}[b]{0.485\linewidth}
\centering
\includegraphics[scale=0.30]{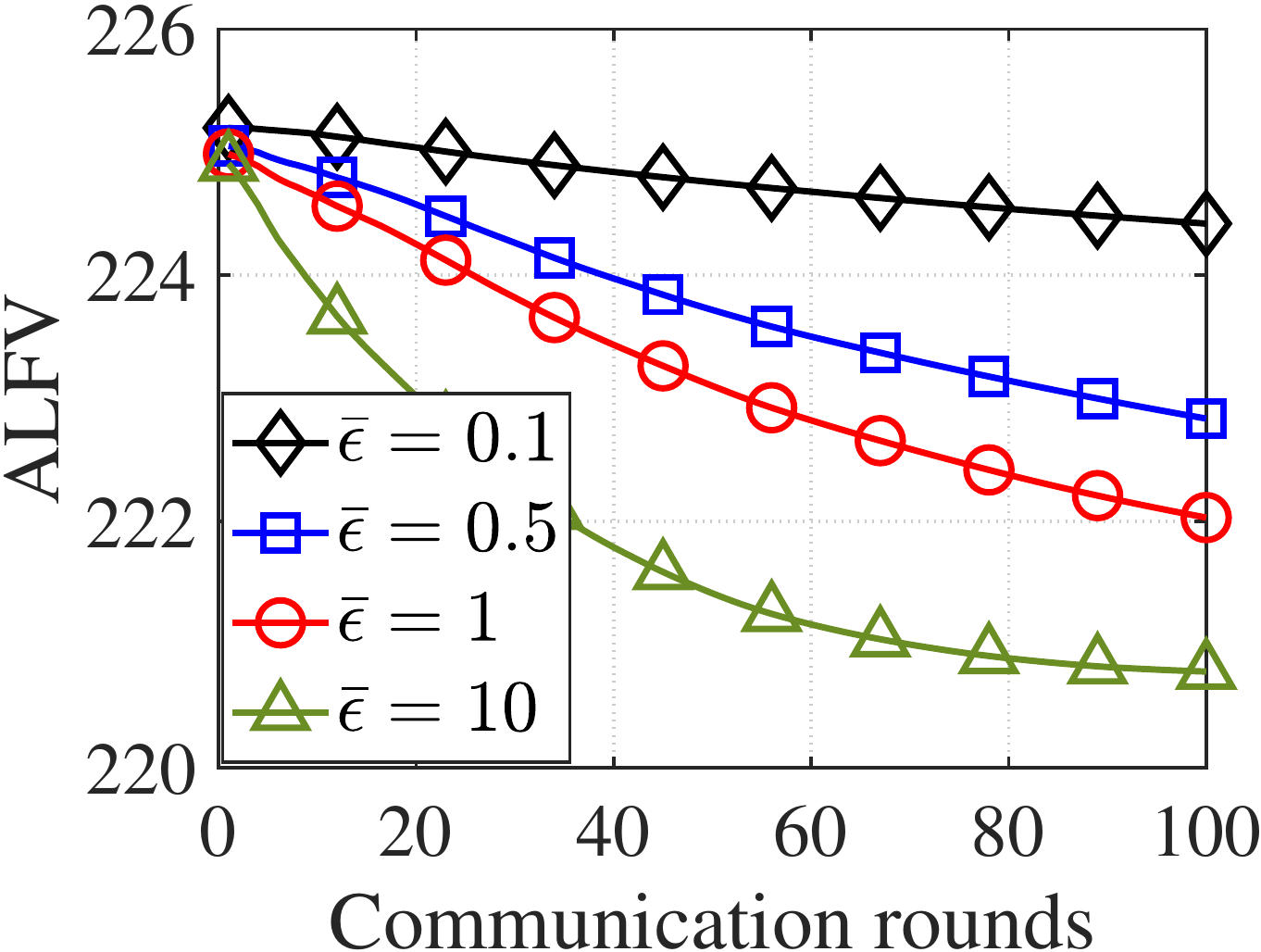} \label{eps_accuracy_b}
\centerline{\scriptsize{(b)}}\medskip
\end{minipage}
\vspace{-0.15cm}
\caption{(a) Testing accuracy and (b) ALFV  versus communication rounds, of the proposed FedSPD-DP algorithm for four different total privacy loss values. }
\vspace{-0.15cm}
\label{eps_accuracy}
\end{figure}

Figures \ref{eps_accuracy}(a) and \ref{eps_accuracy}(b) respectively depict  testing accuracy and ALFV versus communication rounds for various values of total privacy loss $\bar{\epsilon}$ under the system parameters of  $K=20$, $Q = 5$, $b=10$.  One can see from this figure that the testing accuracy is higher for larger total privacy loss $\bar{\epsilon}$, and as expected,  ALFV  decreases with communication round.  The simulation results are consistent with Remark \ref{remark:remark_impact_of_DP}, mentioning that stronger privacy protection (smaller $\epsilon$) will downgrade the learning performance. Hence, there exists a tradeoff between learning performance and the total privacy loss.


\vspace{-0.3cm}
\subsection{Impact of   $K$ (number of  randomly selected clients)}
\begin{figure}[t!]
\begin{minipage}[b]{0.485\linewidth}
\centering
\includegraphics[scale=0.30]{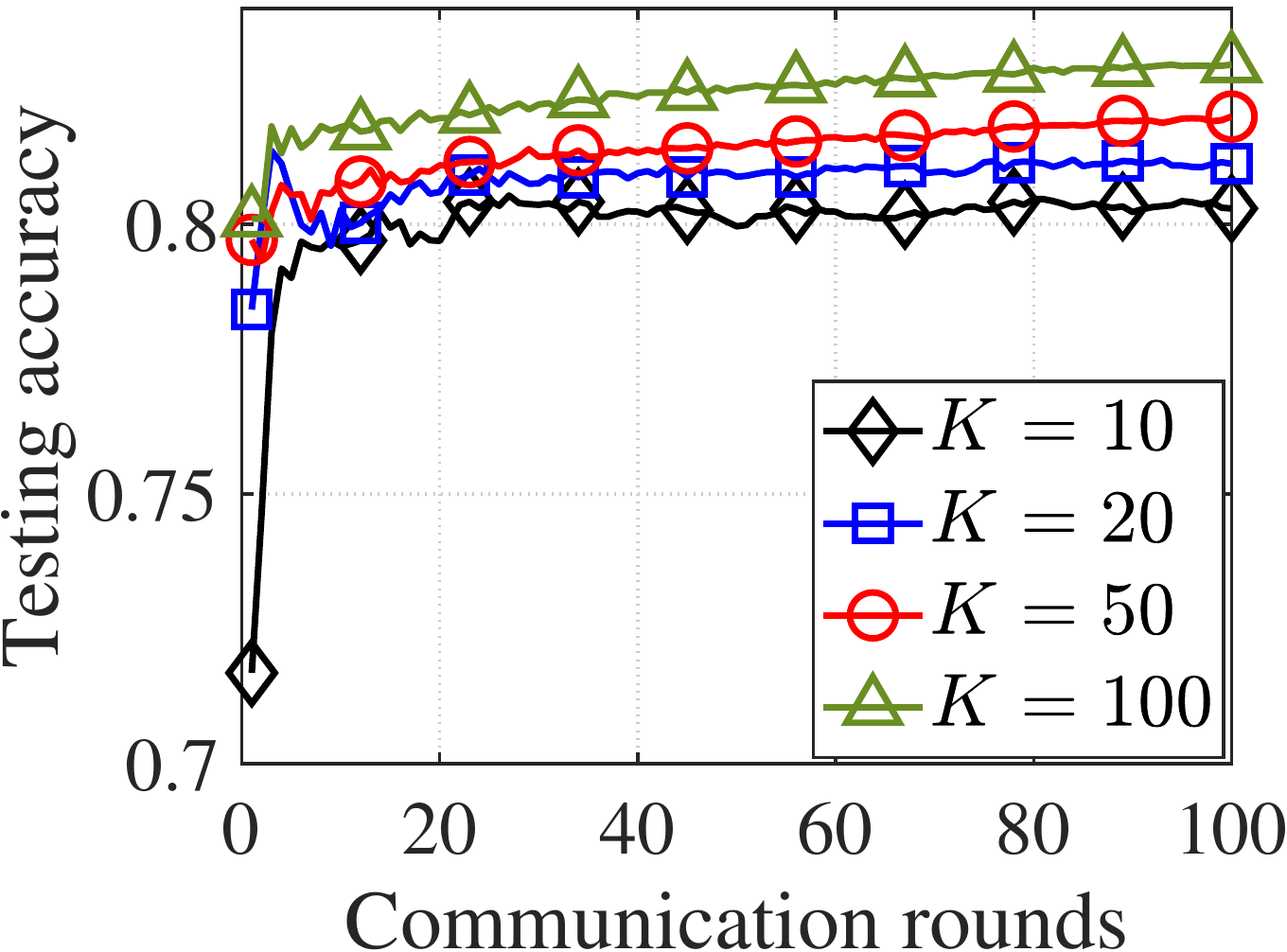}
\centerline{\scriptsize{(a)   }}\medskip
\vspace{-0.15cm}
\end{minipage}
\hfill
\begin{minipage}[b]{0.485\linewidth}
\centering
\includegraphics[scale=0.30]{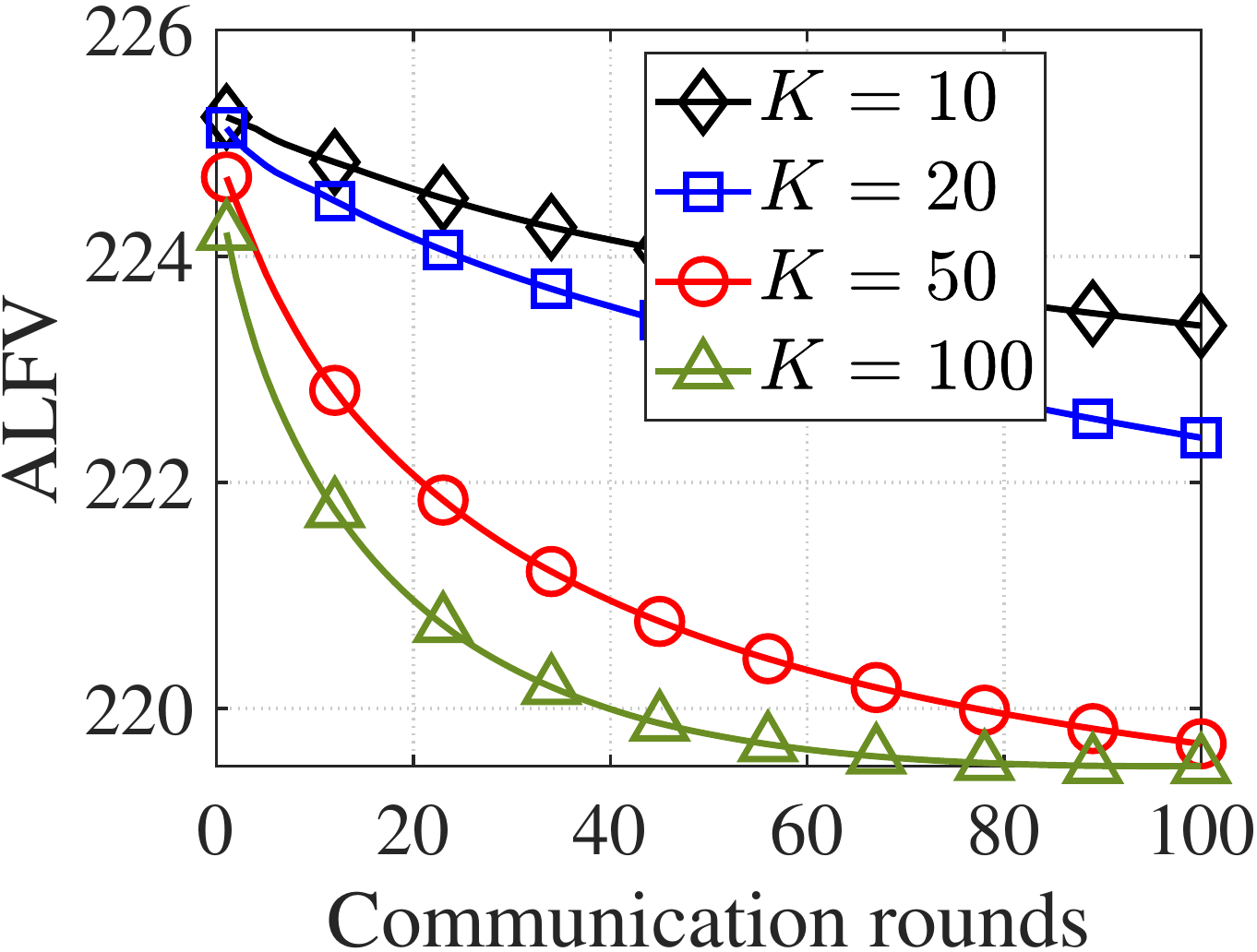}
\centerline{\scriptsize{(b) }}\medskip
\vspace{-0.15cm}
\end{minipage}
\caption{(a) Testing accuracy and (b) ALFV versus communication rounds, of the proposed FedSPD-DP algorithm for four different $K$ values selected through sampling WOR.}
\vspace{-0.15cm}
\label{client_accuracy}
\end{figure}

In this experiment, we show   testing accuracy and ALFV for  various values  $K$ under the system parameters  of $\bar{\epsilon}=1$, $Q=5$, $b=10$.    Figure \ref{client_accuracy} shows that the convergence rate improves as $K$ increases (i.e., larger client sampling probability).  Compared with the FCP (i.e., $K=100$), the PCP introduces randomness in selecting clients, leading to slower convergence.
\subsection{Impact of  $Q$ (number of local SGD steps)}
\begin{figure}[t]
\begin{minipage}[b]{0.485\linewidth}
\centering
\includegraphics[scale=0.30]{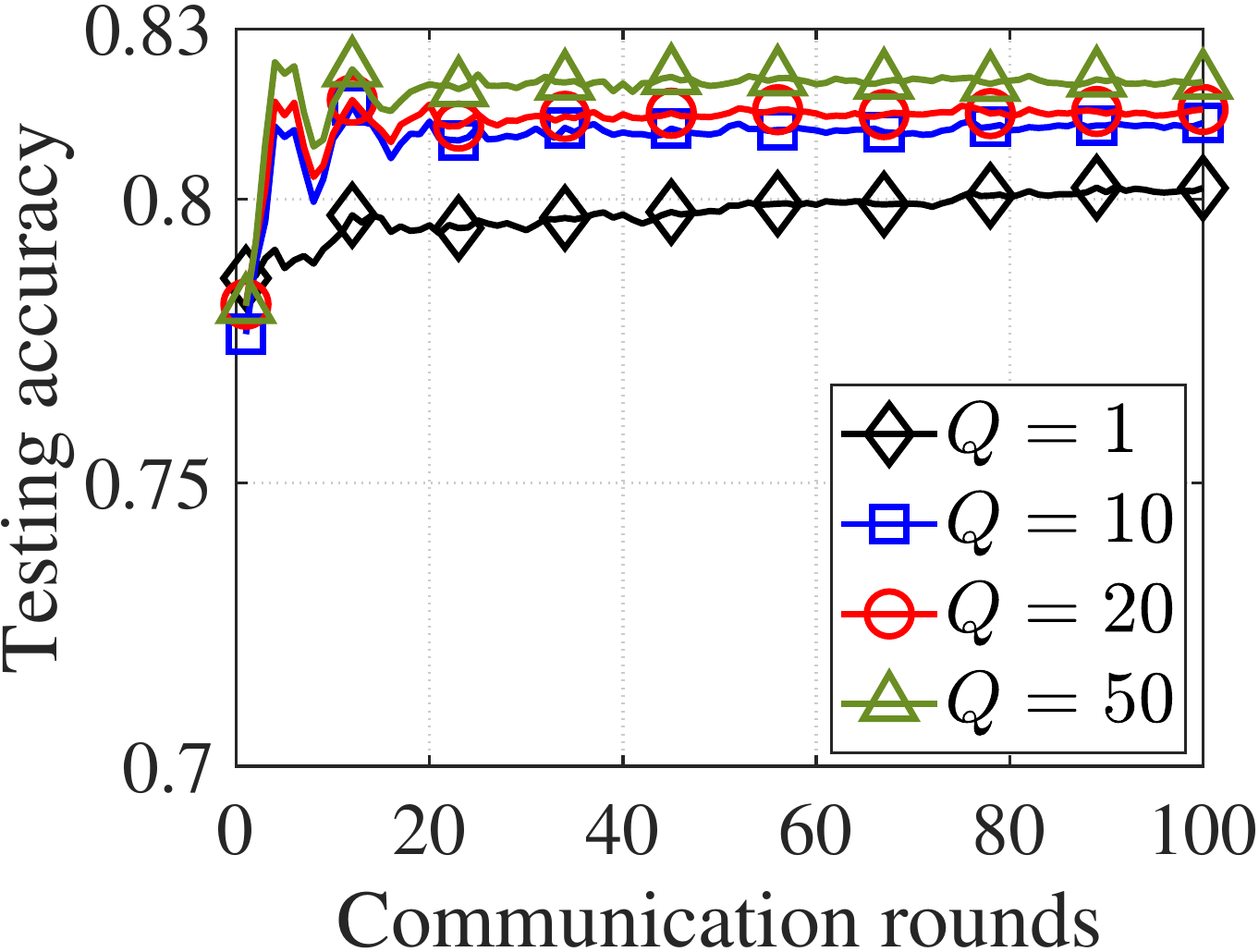}
\centerline{\scriptsize{(a)}}\medskip
\end{minipage}
\hfill
\begin{minipage}[b]{0.485\linewidth}
\centering
\includegraphics[scale=0.30]{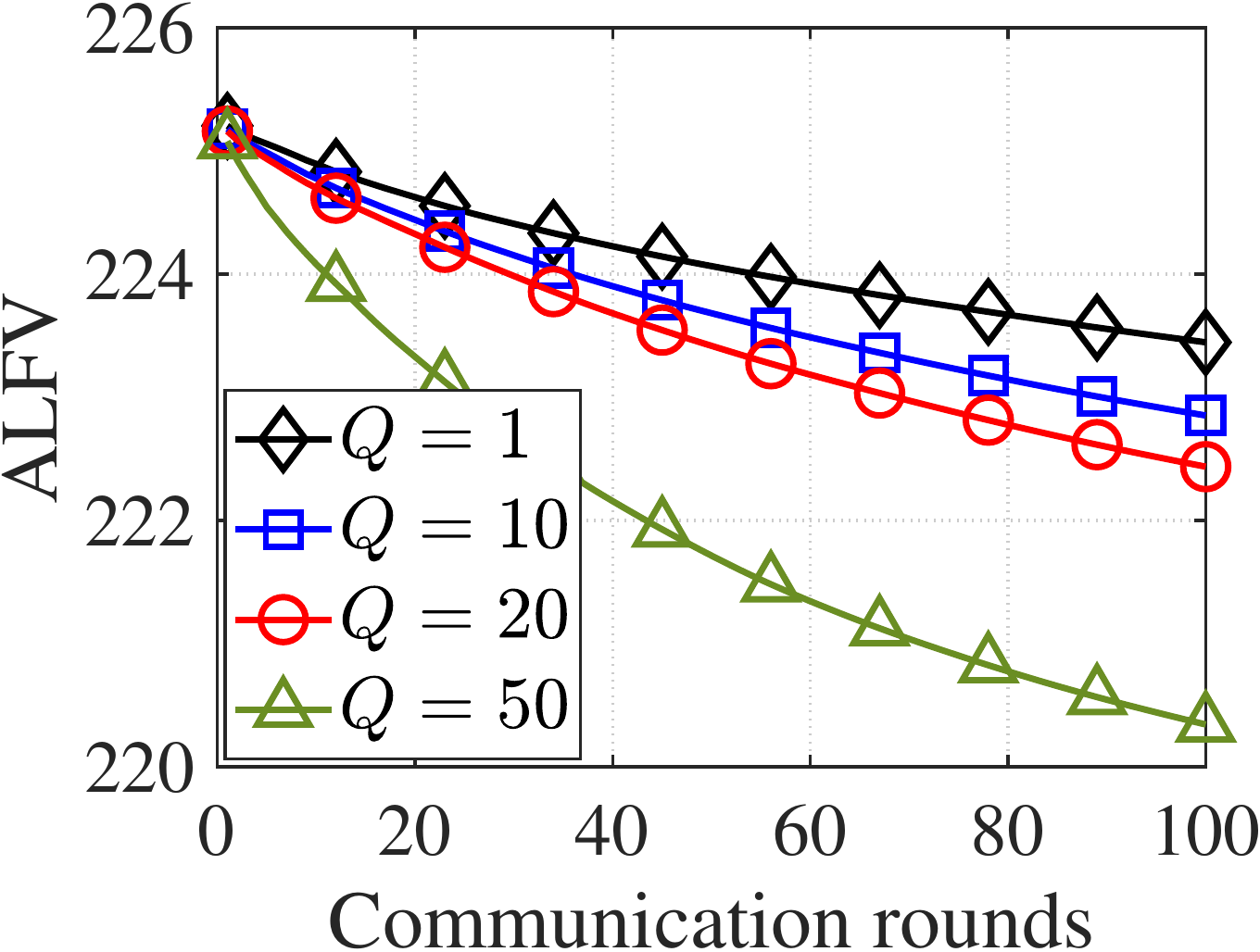}
\centerline{\scriptsize{(b)}}\medskip
\end{minipage}
\vspace{-0.25cm}
\caption{(a) Testing accuracy and (b) ALFV   versus communication rounds, of the proposed FedSPD-DP algorithm for four different $Q$ values.}
\vspace{-0.25cm}
\label{Q_accuracy}
\end{figure}

With the parameters of $\bar{\epsilon}=3$, $K=10$ and $b=10$, Fig. \ref{Q_accuracy} shows that  the learning performance is better for larger $Q$ so is the communication efficiency,  which is consistent with Remark \ref{remark:remark_impact_of_localSGD}.
\subsection{Performance comparison with state-of-the-art  works}   
\begin{figure}[t]
\begin{minipage}[b]{0.485\linewidth}
\centering
\includegraphics[scale=0.30]{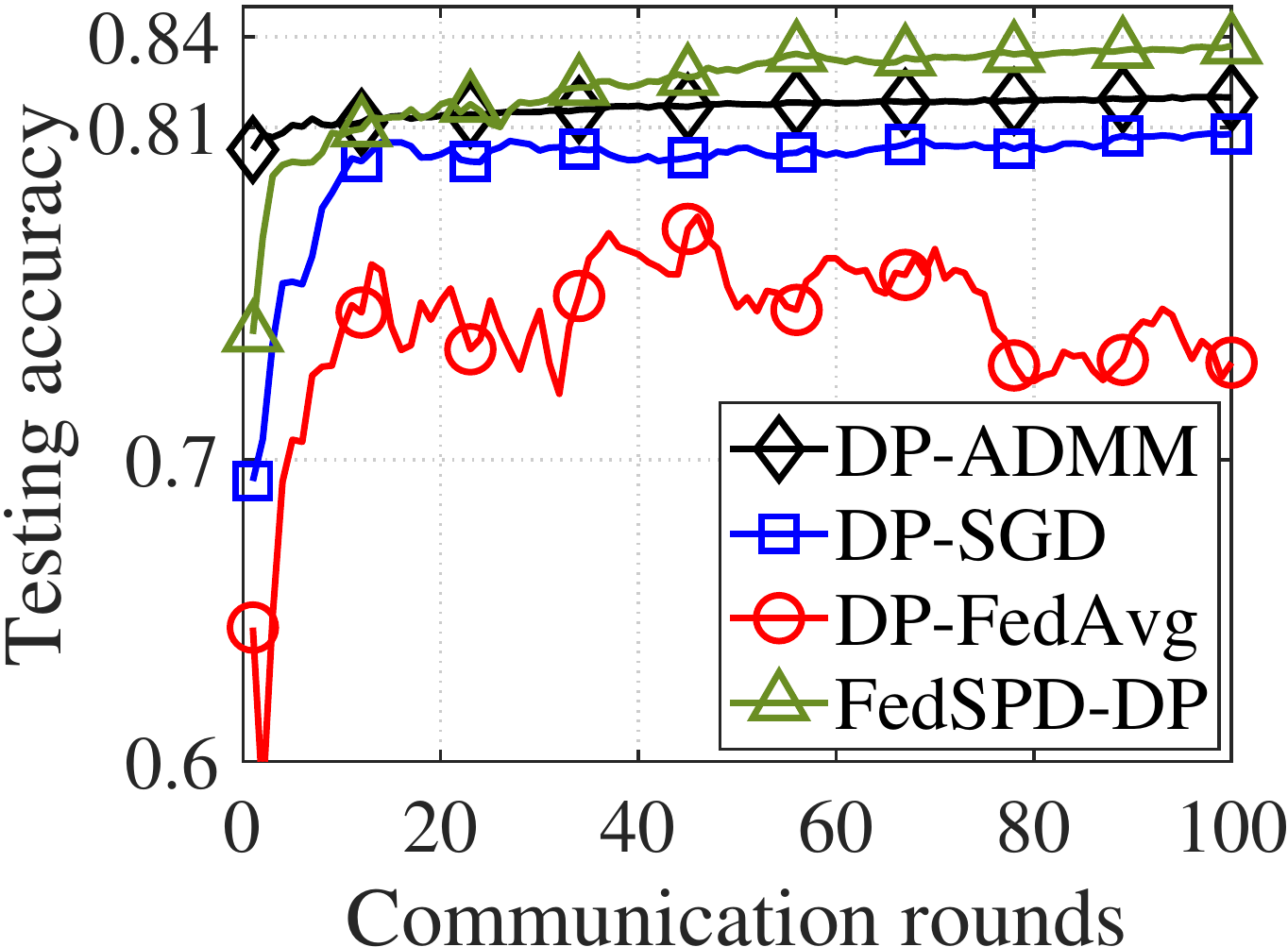}
\centerline{\scriptsize{(a) }}\medskip
\end{minipage}
\hfill
\begin{minipage}[b]{0.485\linewidth}
\centering
\includegraphics[scale=0.30]{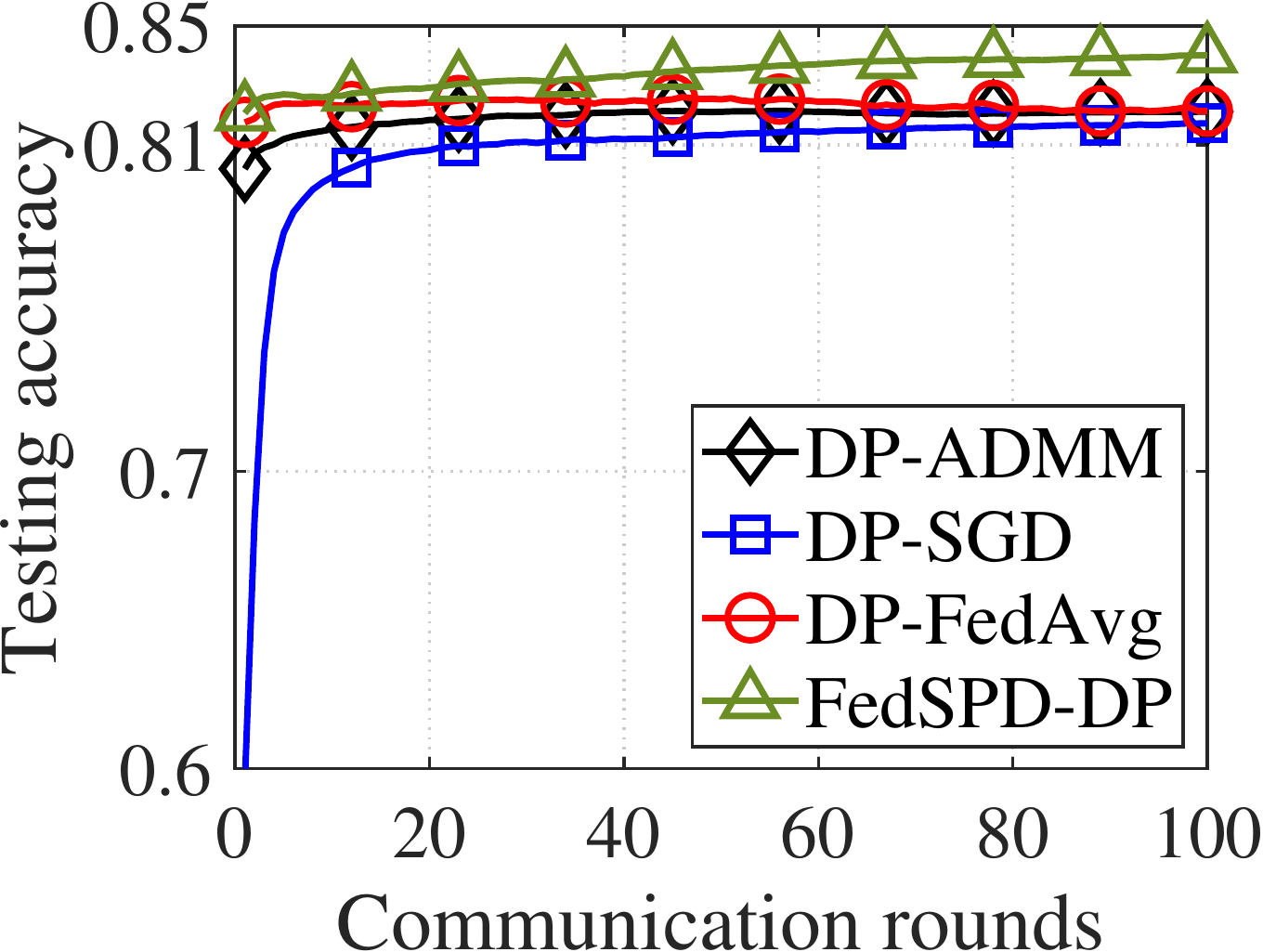}
\centerline{\scriptsize{(b) }}\medskip
\end{minipage}
\hfill
\vspace{-0.15cm}
\caption{Performance comparison of the proposed algorithm with various algorithms for  (a) $\epsilon = 0.1$ and (b) $\epsilon =1$.}
\vspace{-0.15cm}
\label{Compared_with existing works}
\end{figure}

In this experiment, we set the system parameters  $K=20$, $b=10$, $Q=5$.  For a fair comparison, in each round,  $(0.1, 10^{-4})$-DP or $(1, 10^{-4})$-DP is assumed instead of constraining the total privacy loss because some algorithms under comparison do not allow for privacy amplification. For DP-SGD and DP-ADMM   without client sampling in the PS, the  participated clients in each round are the same. It can be observed from   Fig. \ref{Compared_with existing works} that the proposed approach has superior performance over all the other methods  under test for both low and high privacy protection levels. Moreover, we notice that the performance of DP-FedAvg is considerably downgraded when privacy protection  is strong (i.e., $\epsilon$ is small).  The proposed algorithm outperforms DP-ADMM because: $(i)$ DP-ADMM performs the subgradient descent at each communication round, while it can not guarantee the optimal solution for the non-smooth problems due to the occurrence of multiple subgradients; $(ii)$ the proposed algorithm makes use of multiple steps of local SGD in each communication round, thereby yielding better performance.  Finally, the reason that our algorithm performs better than DP-SGD algorithm are that the latter performs gradient clipping at each local update, and consequently, some gradient information gets lost, resulting in worse learning performance.

\begin{figure}[t]
\begin{minipage}[b]{0.485\linewidth}
\centering
\includegraphics[scale=0.30]{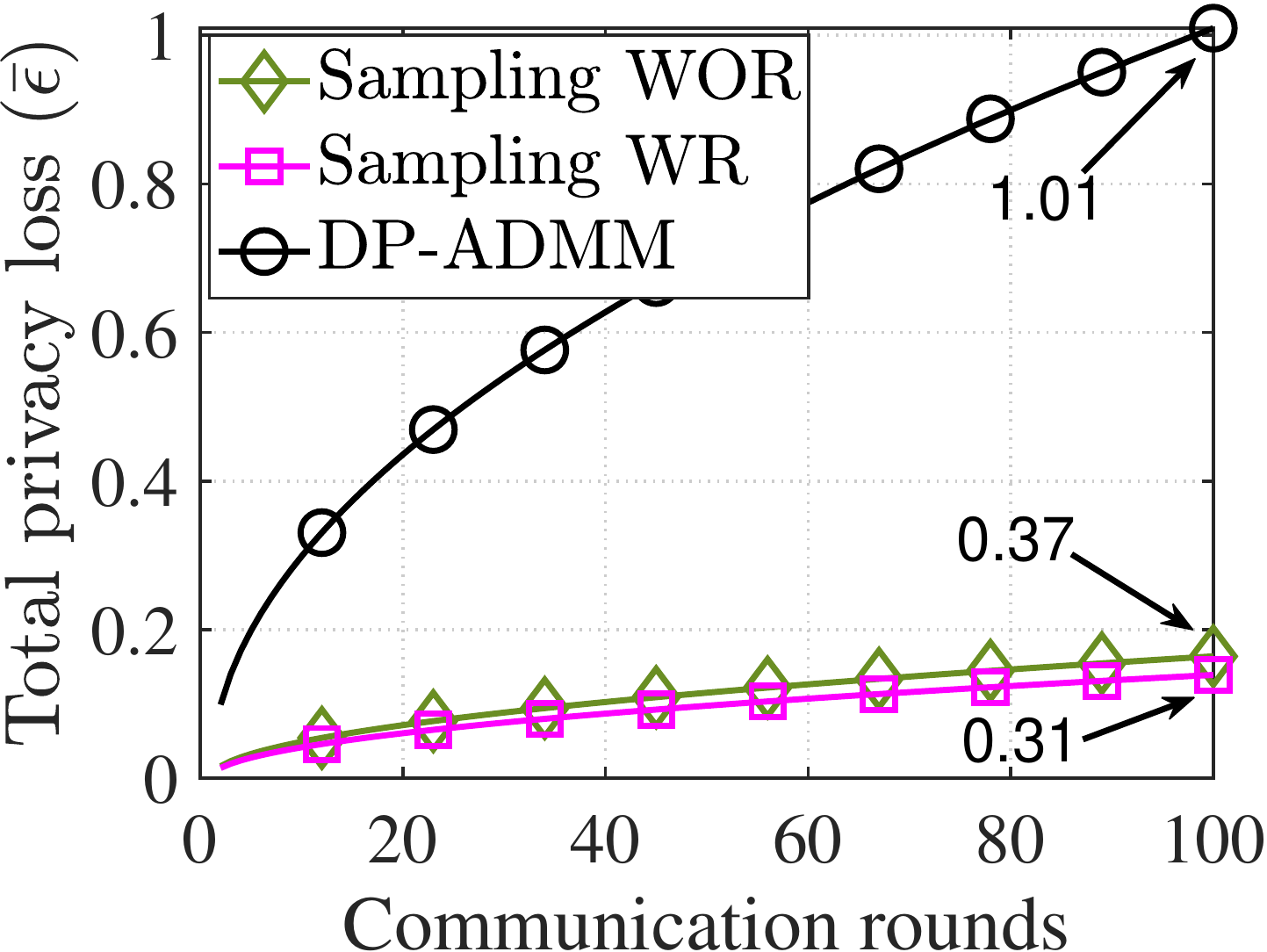}
\centerline{\scriptsize{(a)   }}\medskip
\end{minipage}
\hfill
\centering
\begin{minipage}[b]{0.485\linewidth}
\includegraphics[scale=0.30]{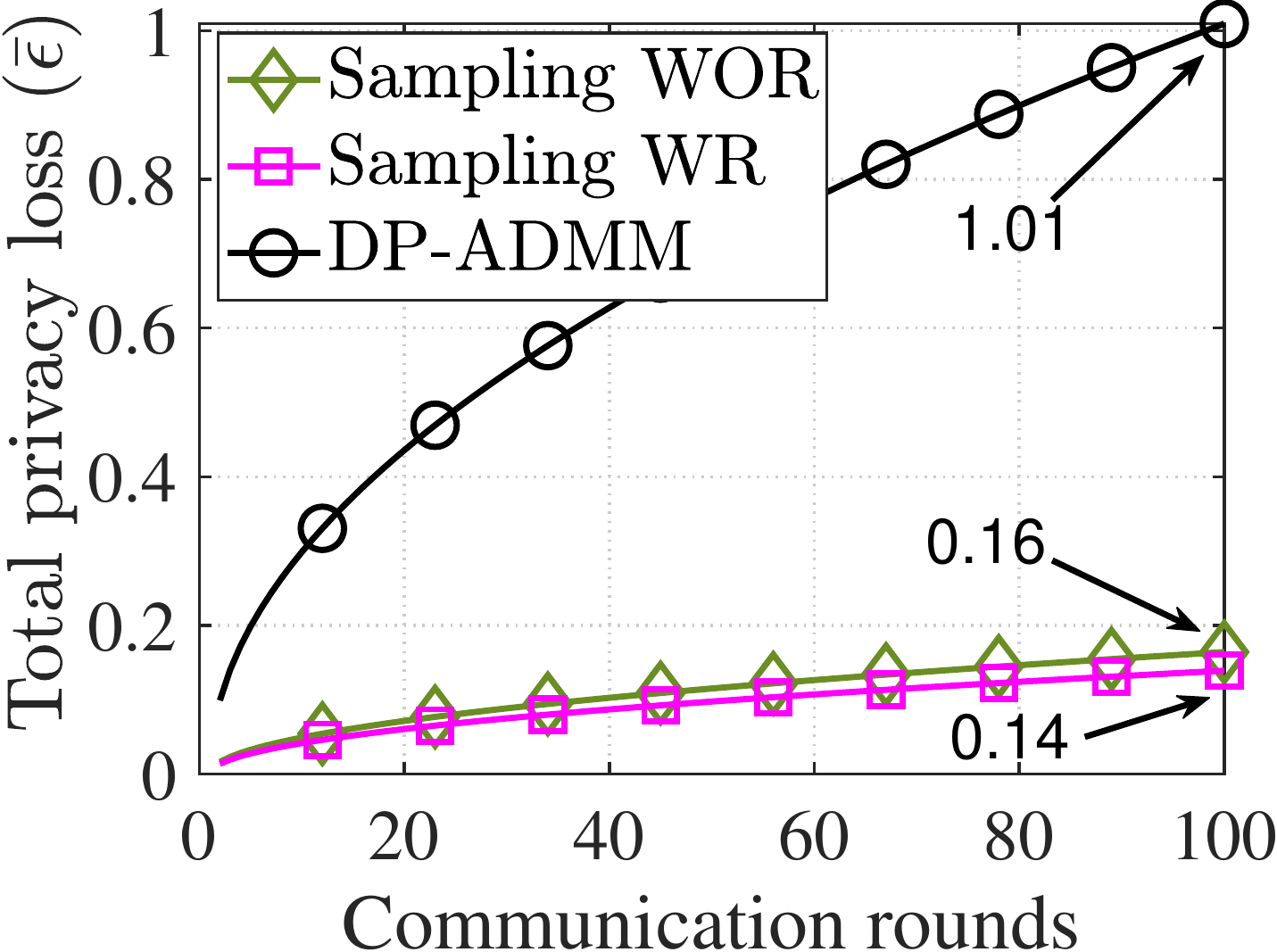}
\centerline{\scriptsize{(b)   }}\medskip
\end{minipage}
\vspace{-0.15cm}
\caption{Comparison of the proposed FedSPD-DP and DP-ADMM  in terms of total privacy loss  for (a) FCP and (b) PCP with data sampling WOR and  WR. }
\vspace{-0.25cm}
\label{Comparison_total_privacy}
\end{figure}

To further show the impact of local SGD and the PCP on  total privacy loss, we  compare  the amounts of the total privacy loss of  DP-ADMM and the proposed algorithm under FCP and PCP with   $(0.1, 10^{-4})$-DP at each round. As shown in Fig. \ref{Comparison_total_privacy}, the total privacy loss in our approach is significantly smaller than in DP-ADMM for both FCP and PCP.   The reason for this is that mini-batch SGD in our algorithm brings privacy amplification as demonstrated in \cite{li2020secure} on one hand, and the noise power $\sigma_{i,t}^2$ required to guarantee the same DP level for our algorithm is considerably smaller than that for DP-ADMM under the privacy loss constraint on the other hand.  From  Fig. \ref{Comparison_total_privacy}(b), one can observe that the total privacy loss under PCP is much smaller than under FCP, which is consistent with  Remark \ref{remark:remark_impact_of_PCP}. Furthermore, the experiment results show  that the  total privacy loss under data sampling WR is  slightly smaller than data sampling WOR as stated in Remark \ref{remark:remark_dp}.
%
\section{Conclusion}\label{sec: conclusion}
We have presented an FL algorithm, i.e., the FedSPD-DP algorithm, which is based on a new primal-dual optimization framework, using DP (for privacy protection), multiple local SGD steps in each communication round (for communication efficiency), and the PCP (for handling
the straggler effect). The learning performance of the proposed FedSPD-DP algorithm has been analyzed in Theorem \ref{Thm:total_budget} and Theorem \ref{Thm:dp theorem2}. The analytical results from the two theorems unveil insightful perspectives and impact of various system parameters in the FL system on the learning performance of the proposed algorithm including the total performance loss. Finally, some experimental results were provided to justify the efficacy of the proposed algorithm, together with its superior performance over some   state-of-the-art algorithms.

\begin{appendices}
\section{Proof of Lemma \ref{lemma:sensitivity}}\label{appdix: proof of Lemma}
\textit{Proof:}  Assume  $\mathcal{D}_{i}$ and $\mathcal{D}_{i}^{\prime}$ are the two neighboring datasets.   According to \eqref{eqn:global sensitivity_f}, the sensitivity of ${\bm y}_{i}^{t}$ is given as follows,
\vspace{-0.2cm}
\begin{small}
\begin{align} \label{eqn:sensitivity_y}
s_{i,t} =& \max _{\mathcal{D}_{i}, \mathcal{D}_{i}^{\prime}} \Big\|{\bm y}_{i, \mathcal{D}_{i}}^{t} -  {\bm y}_{i, \mathcal{D}_{i}^{\prime}}^{t}  \Big\| \notag \\
=& \max _{\mathcal{D}_{i}, \mathcal{D}_{i}^{\prime}} \Big\|\xb_{i, \mathcal{D}_{i}}^{t} -  \xb_{i, \mathcal{D}_{i}^{\prime}}^{t} - \frac{1}{\rho} \big(  \lambdab_{i, \mathcal{D}_{i}}^{t} -  \lambdab_{i, \mathcal{D}_{i}^{\prime}}^{t}\big)  \Big\| \notag \\
\overset{(a)}{=}&   \max _{\mathcal{D}_{i}, \mathcal{D}_{i}^{\prime}} \Big\|\xb_{i, \mathcal{D}_{i}}^{t} -  \xb_{i, \mathcal{D}_{i}^{\prime}}^{t} - \frac{1}{\rho} \big(  \lambdab_{i}^{t-1} -  \lambdab_{i}^{t-1}\big)   \notag\\
& +    \xb_{i, \mathcal{D}_{i}}^{t} -  \xb_{i, \mathcal{D}_{i}^{\prime}}^{t} + \big(   \xb_{0}^{t} - \xb_{0}^{t} \big)  \Big\|  \notag\\
=&  2 \max _{\mathcal{D}_{i}, \mathcal{D}_{i}^{\prime}} \big\|\xb_{i, \mathcal{D}_{i}}^{t} -  \xb_{i, \mathcal{D}_{i}^{\prime}}^{t}   \big\|,
\end{align}
\end{small}

\vspace{-0.3cm}
\noindent where  $(a)$ follows  from  $\lambdab_{i}^{t} = \lambdab_{i}^{t-1}- \rho\big( \xb_{i}^{t}-\xb_{0}^{t}\big)$  and  the fact that $\mathcal{D}_{i}$ and $\mathcal{D}_{i}^{\prime}$ are neighboring datasets.  By plugging \eqref{eqn: local update_primal_2}  into \eqref{eqn:sensitivity_y},  we have
\begin{small}
\begin{align}
&\big\| \xb_{i, \mathcal{D}_{i}}^{t} - \xb_{i, \mathcal{D}_{i}^{\prime}}^{t} \big\|  =  \frac{1}{Q} \Big\| \sum_{r=0}^{Q-1} \big(\xb_{i, \mathcal{D}_{i}}^{t,r} - \xb_{i, \mathcal{D}_{i}^{\prime}}^{t,r} \big) \Big\| \notag \\
= & \frac{1}{Q} \sum_{r=0}^{Q-1} \Big\|    \text{prox}_{ (\gamma_i^t + \rho )^{-1}   R_{i} }  \Big(  (\gamma_i^t + \rho )^{-1} \big(\gamma_i^t \xb_{i, \mathcal{D}_{i}}^{t,r-1} + \rho \xb_{0}^{t}  +   \lambdab_{i}^{t-1}  \notag \\
& -   \nabla f_{i}(\xb_{i, \mathcal{D}_{i}}^{t,r-1}, B_{i}^{t,r})\big)\Big) -   \text{prox}_{ (\gamma_i^t + \rho )^{-1}   R_{i} }  \Big(  (\gamma_i^t + \rho )^{-1}  \notag \\
& \cdot \big( \gamma_i^t \xb_{i, \mathcal{D}_{i}^{\prime}}^{t,r-1} + \rho \xb_{0}^{t}  +   \lambdab_{i}^{t-1} -   \nabla f_{i}(\xb_{i, \mathcal{D}_{i}^{\prime}}^{t,r-1}; (B_{i}^{t,r})^{\prime}) \big)\Big)\Big\| \notag\\
\overset{(b)}{\leq} & \frac{1}{ Q(\rho +\gamma_i^t)}  \sum_{r=0}^{Q-1} \Big\|    \nabla f_{i}(\xb_{i, \mathcal{D}_{i}}^{t, r-1}; B_{i}^{t,r}) - \nabla f_{i}(\xb_{i, \mathcal{D}_{i}^{\prime}}^{t, r-1}; (B_{i}^{t,r})^{\prime} )   \Big\| \notag \\
& +   \frac{\gamma_i^t}{ Q(\rho + \gamma_i^t)}  \sum_{r=0}^{Q-1} \Big\|  \xb_{i, \mathcal{D}_{i}}^{t,r-1} - \xb_{i, \mathcal{D}_{i}^{\prime}}^{t,r-1} \Big\|,  \label{eqn:sensitivity of x}
\end{align}
\end{small}

\vspace{-0.2cm}
\noindent where $(b)$ follows from the nonexpansive property  of proximal operater \cite{parikh2014proximal}, i.e., $\|\operatorname{prox}_{\alpha R} (\xb) - \operatorname{prox}_{\alpha R} (\mathbf{y}) \| \leq \alpha \| \xb - \mathbf{y} \|$.  $\nabla f_{i} (\xb_{i, \mathcal{D}_{i} }^{t, r-1}; B_{i}^{t,r})$ and $ \nabla f_{i} (\xb_{i, \mathcal{D}_{i}^{\prime}}^{t, r-1}; (B_{i}^{t,r})^{\prime})$ denote the  mini-batch gradient  over the mini-batch data $B_{i}^{t,r}$ and $(B_{i}^{t,r})^{\prime}$ that randomly sampled from  $ \mathcal{D}_{i}$ and  $ \mathcal{D}_{i}^{\prime}$, respectively. For notational simplicity, let

\vspace{-0.4cm}
\begin{small}
\begin{align}
\!\!\!g_{B_{i}^{t,r}, (B_{i}^{t,r})^{\prime}}^{t, r-1}  &\triangleq  \Big\|    \nabla f_{i}(\xb_{i, \mathcal{D}_{i} }^{t, r-1}; B_{i}^{t,r}) - \nabla f_{i}(\xb_{i, \mathcal{D}_{i}^{\prime}}^{t, r-1}; (B_{i}^{t,r})^{\prime} )   \Big\| \leq 2G. \label{eqn:gb_notation}
\end{align}
\end{small}

\vspace{-0.3cm}
\noindent Then, by plugging \eqref{eqn:gb_notation} into \eqref{eqn:sensitivity of x}, we have
\begin{small}
\begin{align}
&  \big\|\xb_{i, \mathcal{D}_{i}}^{t}-\xb_{i, \mathcal{D}_{i}^{\prime}}^{t}\big\|    \notag\\
\leq &\sum_{r=0}^{Q-1} \frac{g_{B_{i}^{t,r}, (B_{i}^{t,r})^{\prime}}^{t, r-1} }{ Q(\rho +\gamma_i^t)} +    \frac{\gamma_i^t}{ Q (\rho + \gamma_i^t)}  \sum_{r=0}^{Q-1} \Big\|  \xb_{i, \mathcal{D}_{i}}^{t,r-1} - \xb_{i, \mathcal{D}_{i}^{\prime}}^{t,r-1} \Big\| \notag \\
\leq & \frac{1 }{ Q(\rho +\gamma_i^t)} \sum_{r=0}^{Q-1} \bigg(g_{B_{i}^{t,r}, (B_{i}^{t,r})^{\prime}}^{t, r-1} +   \frac{\gamma_i^t}{ Q(\rho + \gamma_i^t)} \cdot g_{B_{i}^{t,r-1}, (B_{i}^{t,r-1})^{\prime}}^{t, r-2} \notag \\
& + \cdots +  \big(\frac{\gamma_i^t}{ Q(\rho + \gamma_i^t)} \big)^{r-1} \cdot g_{B_{i}^{t,0}, (B_{i}^{t,1})^{\prime}}^{t, 1}  \notag \\
&+ \bigg( \frac{\gamma_i^t}{ Q(\rho + \gamma_i^t)} \bigg)^{r}     \Big\| \xb_{i, \mathcal{D}_{i}}^{t,0} - \xb_{i, \mathcal{D}_{i}^{\prime}}^{t,0} \Big\| \bigg).  \label{eqn:without replacement}
\end{align}
\end{small}According to Algorithm \ref{alg:ADMM2}, we have  $\xb_{i, \mathcal{D}_{i}}^{t,0} = \xb_{i, \mathcal{D}_{i}}^{t-1,Q-1}$ and $\xb_{i, \mathcal{D}_{i}}^{t-1,Q-1} = \xb_{i, \mathcal{D}_{i}^{\prime}}^{t-1,Q-1}$, which   implies $\sum_{r=0}^{Q-1}  \Big\| \xb_{i, \mathcal{D}_{i}}^{t,0} - \xb_{i, \mathcal{D}_{i}^{\prime}}^{t,0} \Big\| =0$.
Then, by Remark \ref{remark:remark1}, we have
\vspace{-0.2cm}
\begin{small}
\begin{subequations}\label{eqn:gradient_WOR}
\begin{align}
&\sum_{r=0}^{Q-1} g_{B_{i}^{t,r}, (B_{i}^{t,r})^{\prime}}^{t, r-1} \leq  \sum_{r=0}^{Q-1}  2G,  \\
&\sum_{r=0}^{Q-1}\frac{\gamma_i^t}{ Q(\rho + \gamma_i^t)} \cdot g_{B_{i}^{t,r-1}, (B_{i}^{t,r-1})^{\prime}}^{t, r-2} \leq \sum_{r=0}^{Q-1}  \frac{2G }{ Q },   \\
&~~~~~~~~~ \vdots \notag\\
& \sum_{r=0}^{Q-1} \bigg( \frac{\gamma_i^t}{ Q(\rho + \gamma_i^t)} \bigg)^{r-1}   g_{B_{i}^{t,1}, (B_{i}^{t,1})^{\prime}}^{t, 0}  \leq \sum_{r=0}^{Q-1} 2G \Big( \frac{1}{Q} \Big)^{r-1}.
\end{align}
\end{subequations}
\end{small}By  plugging  \eqref{eqn:gradient_WOR} into \eqref{eqn:without replacement},  when $Q >1$, we obtain
\begin{small}
\begin{align}
\big\|\xb_{i, \mathcal{D}_{i}}^{t}-\xb_{i, \mathcal{D}_{i}^{\prime}}^{t}\big\|   \leq & \frac{2G }{ Q(\rho +\gamma_i^t)} \sum_{r=0}^{Q-1} \Big(1 + \frac{1}{Q} + \cdots + (\frac{1}{Q})^{r}  \Big) \notag\\
= &  \frac{2G }{    (\rho +\gamma_i^t)}  \frac{1- (\frac{1}{Q})^{r}}{1-\frac{1}{Q}}  \notag\\
\leq &\frac{2G }{ (\rho +\gamma_i^t)} \frac{Q}{Q-1}. \label{eqn:without replacement_WOR}
\end{align}
\end{small}When $Q=1$,  i.e., each communication round  performs one step of SGD, we have
\begin{small}
\begin{align}
&\big\| \xb_{i, \mathcal{D}_{i}}^{t} - \xb_{i, \mathcal{D}_{i}^{\prime}}^{t} \big\|  \notag \\
\leq & \frac{1}{  (\rho +\gamma_i^t)}    \Big\|    \nabla f_{i}(\xb_{i}^{t, 0}; B_{i}^{t,1}) - \nabla f_{i}(\xb_{i}^{t, 0}; (B_{i}^{t,1})^{\prime} )\Big\|   \notag\\
\leq &  \frac{2G}{ (\rho +\gamma_i^t)}. \label{eqn:sensitivity of WR(Q=1)}
\end{align}
\end{small}By \eqref{eqn:sensitivity_y}, \eqref{eqn:without replacement_WOR}  and \eqref{eqn:sensitivity of WR(Q=1)}, we obtain   Lemma \ref{lemma:sensitivity}. $\hfill\blacksquare$

\section{Proof of Theorem \ref{Thm:total_budget}} \label{subsec:the_Proof_of_Theorem_dp}
We first compute the total privacy loss using the moment accountant method in \cite{abadi2016deep,huang2019dp}. Note that, the works in \cite{abadi2016deep} and \cite{huang2019dp} did not consider data sampling strategies, PCP and local SGD. To make the paper self-contained, we present the details following the some ideas as in  \cite{huang2019dp}.  Let $u_{i}$  be the unique data sample between   neighboring datasets $\mathcal{D}_{i}$ and $\mathcal{D}_{i}^{\prime}$ and define the $\tau$-th log moment of the privacy loss of client $i$ at the $t$-th communication round as follows
\begin{small}
\begin{equation}
\begin{aligned}\label{eqn:log_moment}
\alpha_{i}^{t}(\tau) =&   \ln \Big( \mathbb{E}_{\tilde{{\bm y}}_{i}^{t}} \Big[ \Big( \frac{\operatorname{Pr}\big[\tilde{{\bm y}}_{i}^{t} \mid \mathcal{D}_{i}\big]}{\operatorname{Pr}\big[ \tilde{{\bm y}}_{i}^{t} \mid \mathcal{D}_{i}^{\prime}\big]} \Big)^{\tau} \Big] \Big).
\end{aligned}
\end{equation}
\end{small}
By \cite[Theorem 2]{huang2019dp}, $\alpha_{i}^{t}(\tau)$ can be derived as,
\begin{small}
\begin{align}
\alpha_{i}^{t}(\tau) =  \frac{q_i^{2}}{1-q_i}\cdot \frac{\tau(\tau+1) \epsilon^{2}}{4 \ln (1.25 / \delta)},
\end{align}
\end{small}where  $q_{i} $ is the  probability  of the $u_i$ being sampled at the $t$-th round,  which is given in \eqref{eqn: q_1}.  Then,   the  log moment of the  total privacy loss of client $i$ after $T$ rounds is obtained by   Theorem  2  (linear composability) in  \cite{abadi2016deep}.

\vspace{-0.5cm}
\begin{small}
\begin{equation}
\begin{aligned} \label{eqn:total_momentaccount}
\alpha_{i}(\tau) &= \E \Big[ \sum_{t=1}^{T} \mathbb{I}(i \in \Sc_t)\alpha_{i}^{t}(\tau)  \Big] = \sum_{t=1}^{T} \mathbb{P} (i \in \Sc_t)\alpha_{i}^{t}(\tau) \\
&=   \frac{ p_i T q_i^{2}   \tau(\tau+1) \epsilon^{2}}{4 (1-q_i) \ln (1.25 / \delta)},
\end{aligned}
\end{equation}
\end{small}where $p_i$ is the probability of client $i$ being selected. Since each round guarantees $(\epsilon, \delta)$-DP, by Theorem 2, we have
\begin{small}
\begin{align}
\ln (\delta) &=\min _{\tau \in \mathbb{Z}^{+}} \big(\alpha_{i}(\tau)-\tau \bar{\epsilon}_{i} \big) \notag\\
&=\min _{\tau \in \mathbb{Z}^{+}} \big(  \frac{p_i T q_i^{2}  \tau(\tau+1) \epsilon^{2}}{4 (1-q_i) \ln (1.25 / \delta)}-\tau \bar{\epsilon}_{i} \big).  \label{eqn:tail_bound}
\end{align}
\end{small}Since $\delta \in (0,1)$  is a small value,  when $\tau = 1$,  \eqref{eqn:tail_bound} implies
\begin{small}
\begin{align}
\frac{p_i T q_i^{2}  \epsilon^{2}}{2 (1-q_i) \ln (1.25 / \delta)}- \bar{\epsilon}_{i} <0. \label{eqn:tau_1}
\end{align}
\end{small}
To solve   \eqref{eqn:tail_bound}, we relax the RHS of \eqref{eqn:tail_bound}  as problem $(\mathcal{P}3)$,
\begin{small}
\begin{subequations} \label{eqn:minxwithtau}
\begin{align}
(\mathcal{P}3) \ \min_{x}~ &  h(x) \triangleq \frac{p_i T q_i^{2} \epsilon^{2}  x (x+1)}{4 (1-q_{i}) \ln (1.25 / \delta)}- x \bar{\epsilon}_{i}  \label{eqn:minxwithtau_a} \\
\rm {s.t.}~ & x > 0.
\end{align}
\end{subequations}
\end{small}Note that, \eqref{eqn:minxwithtau_a} is a quadratic function of $x$. Thus, \eqref{eqn:minxwithtau} can  be readily solved by letting $ \nabla_x h(x) =0$.  Then, we have
\begin{small}
\begin{align} \label{eqn:minfx}
\!\!\!\!h_{\min}(x) &  \overset{(a)}{\geq}  \frac{3 p_i T q_{i}^{2}   \epsilon^{2}}{16 (1-q_{i}) \ln (1.25 / \delta)} - \frac{\bar{\epsilon}_{i}^{2} (1-q_{i}) \ln (1.25 / \delta)}{p_i T q_{i}^{2}   \epsilon^{2}},
\end{align}
\end{small}where $(a)$ follows from \eqref{eqn:tau_1}. Then, according to \eqref{eqn:tail_bound} and \eqref{eqn:minfx}, we have
\begin{small}
\begin{align}
\ln(\delta) &=\min _{\tau \in \mathbb{Z}^{+}} \big(  \frac{p_i T q^{2}  \tau(\tau+1) \epsilon^{2}}{4 (1-q_{i}) \ln (1.25 / \delta)}-\tau \bar{\epsilon}_{i} \big) \notag \\
& \geq \frac{3 p_i T q_{i}^{2}   \epsilon^{2}}{16 (1-q_{i}) \ln (1.25 / \delta)} - \frac{\bar{\epsilon}_{i}^{2} (1-q_{i}) \ln (1.25 / \delta)}{p_i T q_{i}^{2}   \epsilon^{2}}, \label{eqn:minlntau}
\end{align}
\end{small}which further implies
\begin{small}
\begin{align}
\ln(1/\delta)  &\leq -\frac{3 p_i T q_{i}^{2}   \epsilon^{2}}{16 (1-q_{i}) \ln (1.25 / \delta)} + \frac{\bar{\epsilon}_{i}^{2} (1-q_{i}) \ln (1.25 / \delta)}{p_i T q_{i}^{2}   \epsilon^{2}} \notag \\
&\leq   \frac{\bar{\epsilon}_{i}^{2} (1-q_{i}) \ln (1.25 / \delta)}{p_i T q_{i}^{2}   \epsilon^{2}}. \label{eqn:minlntau_inverse}
\end{align}
\end{small}
By rearranging  \eqref{eqn:minlntau_inverse}, we have
\begin{small}
\begin{equation}
\begin{aligned}
\bar{\epsilon}_{i} \geq   \frac{q_i}{\sqrt{1-q_i}} \sqrt{ \frac{p_i T  \ln(1/\delta)}{ \ln (1.25 / \delta)}} \epsilon.
\end{aligned}
\end{equation}
\end{small}

\vspace{-0.25cm}
\noindent Therefore, there exists a constant $ c_{0}$ such that the total privacy loss of client $i$ after $T$  rounds  satisfies:
\begin{small}
\begin{align} \label{eqn:total_privacyloss_appdix}
\bar{\epsilon}_{i} =c_{0} \frac{q_i \sqrt{p_i}}{\sqrt{1-q_i}} \sqrt{T} \epsilon, \ \forall  i \in [N].
\end{align}
\end{small}
Thus, Theorem   \ref{Thm:total_budget} is proved.   $\hfill\blacksquare$

\section{Proof of Theorem \ref{Thm:dp theorem2}} \label{appdix: proof of thm 2}

\subsection{Preliminaries}
Before diving into the proof, let us introduce some useful variables and notations for ease of analysis. Specifically, we recapitulated the primal and dual variables to have the following new variables:
\begin{small}
\begin{align} \label{eqn: omega_defs}
& \wbb_i^t \triangleq\left[\begin{array}{c}
\xb_i^t  \\
\xb_0^t  \\
\lambdab_{i}^t
\end{array}\right], ~
~\wbb_{i}^{t,r} \triangleq \left[\begin{array}{c}
	\xb_{i}^{t,r} \\
	\xb_{0}^{t} \\
	\lambdab_{i}^{t}
\end{array}\right],
~\wt \wbb_{i} \triangleq \left[\begin{array}{c}
\xb_i^{\star} \\
\xb_0^{\star}\\
\lambdab_{i}
\end{array}\right], \notag\\
& \wbb_{i}^{t}\triangleq  \frac{1}{Q} \sum_{r=0}^{Q-1} \wbb_{i}^{t,r}, ~\bar \wbb_{i}^{T+1}\triangleq  \frac{1}{T+1}  \sum_{t=0}^{T} \wbb_{i}^{t},
\end{align}
\end{small}where $\xb_i^\star= \xb_0^\star, \forall i$ is the optimal solution to problem ($\mathcal{P}2$) and $\lambdab_{i}$ is the dual variable bounded within $\mathcal{B}_{\bf \lambda}$. To deal with the randomness incurred by the scheme of PCP,   we define the virtual sequence $\{\hat \wbb_i^t\}$ by assuming all clients are active at the $t$-th round, in which $\hat \wbb_i^t$ is given by
{\small
\begin{align}
	\hat \wbb_i^t = \big[(\hat \xb_i^t)^\top, ( \xb_0^t)^\top, (\hat\lambdab_{i}^t)^\top \big]^{\top},
\end{align}}with
\begin{small}
\begin{subequations}
\begin{align}
	&\hat \xb_i^{t, 0} = \xb_i^{t-1, Q}, \label{eqn:update_rule_11}\\
	& \hat \xb_i^{t, r} = \arg\min_{\xb_i}  \wt \Lc_{i}(\xb_{i}, \hat \xb_0^t, \lambdab_{i}^{t-1}, \hat \xb_i^{t, r-1}),0 \leq r \leq Q-1, \\
	& \hat \xb_{i}^{t}=\frac{1}{Q}\sum_{r=0}^{Q-1} \hat \xb_i^{t, r}, \label{eqn: def_virtual_xi}\\
	&\hat \lambdab_{i}^t =  \lambdab_{i}^{t-1}- \rho(\hat \xb_{i}^{t}-\hat \xb_{0}^{t}), \label{eqn: def_virtual_dual}
\end{align}
\end{subequations}
\end{small}
where
\begin{small}
\begin{equation}
\begin{aligned}
	&\wt \Lc_{i}\big(\xb_{i}, \hat \xb_0^t, \lambdab_{i}^{t-1}, \hat \xb_i^{t, r-1} \big)   \\
	 = & f_{i}(\hat \xb_i^{t, r-1}) + \big\langle \nabla f_{i} (\hat \xb_i^{t, r-1}; B_i^{t,r}), \xb_i-  \hat \xb_{i}^{t,r-1} \big\rangle     \\
	&  + \frac{\gamma_i^t}{2}\|\xb_i - \hat \xb_i^{t, r-1}\|^{2} -  \big\langle \lambdab_{i}^{t-1}, \xb_i - \hat \xb_0^t \big\rangle    \\
    & + \frac{\rho}{2} \|\xb_{i} - \hat \xb_0^t\|^{2}+    R_{i}(\xb_{i}).
\end{aligned}
\end{equation}
\end{small}
Next, we introduce two auxiliary functions as follows
 \begin{small}
\begin{align}
&\Psi(\wbb_i^t, \wt \wbb_i) \triangleq  \sum_{i=1}^{N} \big( F_{i} (\xb_i^t)- F_{i}(\xb_i^{\star})  +  \big\langle \boldsymbol{w}_{i}^t- \widetilde{\boldsymbol{w}}_{i},  H(\boldsymbol{w}_{i}^t) \big\rangle \big), \label{eqn: def_psi}\\
& J_{i}(\xb_i^t, \xb_0^t, \lambdab_{i}^t) \triangleq  \frac{1}{p_i} \big(F_i(\xb_i^t)+ \langle -\lambdab_{i}^t, \xb_i^t - \xb_0^t \rangle \big), \label{eqn: def_J}
\end{align}
\end{small}
where
{\small \begin{align} \label{eqn:def_H}
&H(\wbb_i^t)\triangleq\left[\begin{array}{c}
	-\lambdab_{i}^t \\
	\lambdab_{i}^t \\
	\xb_i^t -\xb_0^t
\end{array}\right].
\end{align}}
\vspace{-0.4cm}
\subsection{Sketch of the Proof}
The proof of Theorem \ref{Thm:dp theorem2} relies on bounding $\sum_{i = 1}^{N}\E[\Psi(\bar \wbb_i^{T+1}, \wt \wbb_i) \big]$. The idea is motivated by one feature of $\Psi(\wbb_i, \wt \wbb_i)$ that the supremum of $\sum_{i = 1}^{N}\Psi(\bar \wbb_i^{T+1}, \wt \wbb_i)$ with respect to $\lambdab_{i}$ is equivalent to the optimality gap in \eqref{eqn:criterion}, as shown in Lemma \ref{lem_PSI_supermum}. Thus, by following Lemma \ref{lem_PSI_supermum}, it suffices to obtain the bound of $\sum_{i = 1}^{N}\E\big[\Psi(\bar \wbb_i^{T+1}, \wt \wbb_i)\big]$ to achieve the convergence result.
\vspace{-0.1cm}
\begin{Lemma}\label{lem_PSI_supermum}
For  $\bar \wbb_{i}^{T+1}$ from the variable defined in \eqref{eqn: omega_defs}, the following holds true.
\begin{small}
\begin{align}
&\sup_{\lambdab_{i} \in \mathcal{B}_{\bf \lambda} } \sum_{i=1}^{N}  \Psi( \bar \wbb_i^{T+1}, \widetilde{\boldsymbol{w}}_{i}) \notag \\
=&  \sum_{i=1}^{N}  \big( {F_{i}(\bar \xb_i^{T+1} - F_{i}(\xb_i^{\star})} + \beta \|\bar \xb_i^{T+1} - \bar \xb_0^{T+1} \| \big). \label{lem_PSI_supermum_inequality}
\end{align}
\end{small}
\end{Lemma}

\vspace{-0.20cm}
The bound of $\sum_{i = 1}^{N}\E[\Psi(\bar \wbb_i^{T+1}, \wt \wbb_i)]$ can be achieved by the convexity of $\Psi(\wbb_i, \wt \wbb_i)$ and one-round progress analysis of the function $J_{i}(\xb_i^t, \xb_0^t, \lambdab_{i}^t)$ over the iterates $\xb_i^t$ and $\xb_0^t$. In particular, owing to the convexity of $\Psi(\wbb_i, \wt \wbb_i)$, we have
\begin{small}
\begin{align}
\sum_{i = 1}^{N}\Psi(\bar \wbb_i^{T+1}, \wt \wbb_i) \leq \frac{1}{T+1}\sum_{t =1}^{T+1} \sum_{i = 1}^{N}\Psi(\wbb_i^{t-1}, \wt \wbb_i). \label{eqn: formal_b2}
\end{align}
\end{small}
Based on the definition of $J_{i}(\xb_i^t, \xb_0^t, \lambdab_{i}^t)$ and the client sampling in round $t$,  we establish Lemma \ref{lem: J_connect} which shows that   $\sum_{i = 1}^{N} \E \big[\Psi(\wbb_i^{t-1}, \wt \wbb_i) \big]$ is bounded by the one-round progress of $J(\xb_i, \xb_0, \lambdab_{i})$ plus $\sum_{i = 1}^{N}\Psi(\hat \wbb_i^{t}, \wt \wbb_i)$.
\vspace{-0.15cm}
\begin{Lemma}\label{lem: J_connect}
According to the definition of $J(\cdot)$ in \eqref{eqn:J} and the update rule of Algorithm \ref{alg:ADMM2}, it follows that

\vspace{-0.35cm}
\begin{small}
\begin{align}
\sum_{i = 1}^{N}\E \big[\Psi(\wbb_i^{t-1}, \wt \wbb_i) \big]	\leq &\sum_{i=1}^{N}  \E \big[ J_{i}(\xb_i^{t-1},\xb_0^{t-1}, \lambdab_{i}) - J_{i} (\xb_i^t, \xb_0^t, \lambdab_{i}) \big]  \notag\\
&+ \sum_{ i = 1}^{N}\E \big[\Psi(\hat \wbb_i^t, \wt \wbb_i) \big].\label{lem: J_connect_bound}
\end{align}
\end{small}
\end{Lemma}

\vspace{-0.1cm}
Inspired by the strategies in the proof of \cite[Theorem 4.3]{Hong2017StochasticPG},  the bound of $\sum_{i = 1}^{N}\E\big[\Psi(\hat \wbb_i^{t}, \wt \wbb_i)\big]$ can be established by Lemma \ref{lem: psi_virtual_obtain}. Lastly, putting all these pieces together, doing the telescoping sum over the two sides of \eqref{lem: J_connect_bound} and choosing proper values of certain parameters yield the result in Theorem \ref{Thm:dp theorem2}. More details  can be found in Appendix \ref{sec: formal_proof}.
\begin{Lemma}\label{lem: psi_virtual_obtain}
Suppose that all the clients are active at communication round $t$ and  $ \rho \geq \sqrt{\gamma_i^{t}}, \forall t$,  then it follows that
\vspace{-0.1cm}
\begin{small}
	\begin{align}
& \frac{1}{T+1}\sum_{t = 1}^{T+1} \sum_{i=1}^{N} \E\big[ \Psi(\hat \wbb_i^t, \wt \wbb_i)\big]	\notag  \\
\leq & \frac{1}{T+1}\sum_{t = 1}^{T+1} \sum_{i = 1}^{N} \frac{1}{\gamma_i^{t}} \Big( G^{2}  +  \frac{2\phi^2}{b} + 2d_{\lambda}^{2} + \frac{\rho}{2} \E\big[ \| ( \rho + \gamma_i^{t-1})\xi_i^{t-1} \|^{2} \big] \Big)  	\notag \\
& + \frac{1}{(T+1)Q} \sum_{i = 1}^{N} \frac{ \gamma_i^{T+1} + \rho }{2p_i} \E \big[\|  \xb_i^{0,Q}  -   \xb_i^{\star} \|^{2}  - \|  \xb_i^{T+1,Q} -   \xb_i^{\star} \|^{2} \big]\notag  \\
&+  \frac{\rho}{2} \frac{1}{T+1} \sum_{i = 1}^{N}\frac{1}{p_i} \big(\|\xb_i^{0} - \xb_0^{\star}\|^2 - \|\xb_i^{T+1} - \xb_0^{\star}\|^2 \big) \notag    \\
& + \frac{1}{2\rho}\frac{1}{T+1} \sum_{i = 1}^{N}\frac{1}{p_i} \big(\|\lambdab_i^{0} - \lambdab_{i}\|^2 - \|\lambdab_i^{T+1} - \lambdab_i^{\star}\|^2 \big) \notag  \\
& + \frac{1}{(T+1)Q}\sum_{i = 1}^{N} \frac{1}{p_i} \E \big[R_{i}(  \xb_i^{0,Q}) -  R_{i}( \xb_i^{T+1, Q})\big].\label{lem: psi_virtual_obtain_bd}
\end{align}
\end{small}
\end{Lemma}

\subsection{Completing the Proof of Theorem \ref{Thm:dp theorem2}}\label{sec: formal_proof}
Now we proceed to the formal proof of Theorem \ref{Thm:dp theorem2}.   Summing the inequality \eqref{lem: J_connect_bound} in Lemma \ref{lem: J_connect} from $t = 1$ to $T+1$, and dividing it by $T+1$ yields
{\small \begin{align} \label{eqn: formal_b1}
	&\frac{1}{T+1}\sum_{t = 1}^{T+1}\sum_{i = 1}^{N}\E \big[\Psi(\wbb_i^{t-1}, \wt \wbb_i) \big]	\notag \\
	\leq &\frac{1}{T+1}\sum_{i=1}^{N}  \E \big[J_{i}(\xb_i^{0},\xb_0^{0}, \lambdab_{i}) - J_{i} (\xb_i^{T+1}, \xb_0^{T+1}, \lambdab_{i}) \big] \notag \\
	&+ \frac{1}{T+1}\sum_{t = 1}^{T+1}\sum_{ i = 1}^{N}\E \big[\Psi(\hat \wbb_i^t, \wt \wbb_i) \big].
\end{align}}   Then, plugging \eqref{eqn: formal_b2} and \eqref{lem: psi_virtual_obtain_bd} in Lemma \ref{lem: psi_virtual_obtain} into \eqref{eqn: formal_b1} gives
\begin{small}
	\begin{align}\label{eqn:H_bounded_total_1111}
		 &  \sum_{i=1}^{N}\E \big[ \Psi( \bar \wbb_i^{T+1}, \wt \wbb_i) \big]     \leq     \frac{1}{T+1} \sum_{t=1}^{T+1} \sum_{i=1}^{N}  \E \big[\Psi( \wbb_i^{t-1}, \wt \wbb_i) \big] \notag\\
		\leq & \frac{1}{T+1}  \sum_{i=1}^{N}  \E \big[J_{i}(\xb_{i}^{0}, \xb_{0}^{0}, \lambdab_{i} )- J_{i} (\xb_i^{T+1}, \xb_0^{T+1}, \lambdab_{i}) \big]    \notag \\
	&  + \frac{1}{T+1} \sum_{t=1}^{T+1}  \sum_{i=1}^{N}   \frac{1}{\gamma_i^t} \Big( G^{2} + 2d_{\lambda}^{2} +      \frac{2 \phi^{2}}{b}     +   \frac{\rho}{2} \E[\|( \rho + \gamma_i^{t-1}) \xib_{i}^{t-1} \|^{2} ]    \Big)  \notag \\
		&+  \frac{1}{2(T+1) Q}  \sum_{i=1}^{N} \Big(   \frac{\rho Q}{p_{i}}  \| \xb_{0}^{\star} - \xb_{i}^{0}\|^{2}    +   \frac{ \gamma_i^{T+1} + \rho }{p_i}  \|  \xb_{i}^{0,Q} -   \xb_{i}^{\star} \|^{2}  \notag \\
		&+ \frac{Q}{\rho} \frac{1}{p_{i}} \left\|\lambdab_{i} - \lambdab_{i}^{0} \right\|^{2}  +    \frac{2}{p_i} \E \big[ R_{i}(  \xb_i^{0,Q}) -  R_{i}( \xb_i^{T+1, Q}) \big] \Big).
	\end{align}
\end{small}Using  Lemma \ref{lem_PSI_supermum} and Assumption \ref{Ass: Assumption3},  taking the supreme with respect to $\lambdab_{i}$ on both side of the  (\ref{eqn:H_bounded_total_1111}) yields
\begin{small}
	\begin{align}\label{eqn:H_bounded_total_2222}
		&   \sum_{i =1}^{N}   \E[  F_{i}( \bar{\xb}_{i}^{T+1})-F_{i}(\xb_{0}^{\star} )] + \beta \sum_{i =1}^{N} \E \big[\|\bar{\xb}_{i}^{T+1}-\bar{\xb}_{0}^{T+1}\| \big] \notag\\
		\leq &  \frac{1}{T+1} \bigg( \sum_{i=1}^{N}  \sup_{ \lambdab_{i} \in \Bc_{\lambda}} \E \big[ J_{i} (\xb_{i}^{0}, \xb_{0}^{0}, \lambdab_{i}) - J_{i} (\xb_i^{T+1}, \xb_0^{T+1}, \lambdab_{i}) \big] \notag \\
		& + \underbrace{ \sum_{t=1}^{T+1}  \sum_{i=1}^{N}  \frac{1}{\gamma_i^t}\Big( G^{2} + 2d_{\lambda}^{2}     +  \frac{2 \phi^{2}}{b} +   \frac{\rho}{2}\E \big[ \| ( \rho + \gamma_i^{t-1}) \xib_{i}^{t-1} \|^{2} \big] \Big) }_{ \mathcal{A}}  \notag \\
		& +  \frac{1}{2Q}  \sum_{i=1}^{N} \Big(   \frac{ \rho + \gamma_i^{T+1}    }{p_i}   d_{\mathcal{X}}^{2}  + \frac{\rho Q d_{\mathcal{X}}^{2} }{p_{i}}   +  \frac{d_{\lambda}^2}{\rho}   \frac{Q}{p_{i}}      +    \frac{4G_{R}}{p_i} \Big) \bigg).
	\end{align}
\end{small}
When $Q>1$,  we then obtain the required noise scale to guarantee $(\epsilon, \delta)$-DP by \eqref{eqn:noise for with replacement_0}
\begin{small}
	\begin{equation}
		\begin{aligned} \label{eqn:noisepower}
			\E [\| ( \rho + \gamma_i^{t-1} ) \boldsymbol{\xi}_{i}^{t-1} \|^{2}] =   \frac{32 d G^{2} \ln (1.25/\delta)}{ (Q-1)^{2}  \epsilon^{2} }.
		\end{aligned}
	\end{equation}
\end{small}
By setting $\gamma^{t}  =  \frac{2\sqrt{Q C(Q,\epsilon)     p_i} }{d_{\mathcal{X}}} \sqrt{t}$, where
\begin{small}
\begin{align}\label{eqn: C_f_A}
C(Q,\epsilon)     =  G^{2}  + 2d_{\lambda}^{2} +  \frac{2\phi^{2}}{b}  + \frac{  16 \rho d G^{2} \ln (1.25/\delta)}{  (Q-1)^{2}  \epsilon^{2}}.
\end{align}
\end{small}
Then,  dividing term $\mathcal{A}$ by $T+1$, we have
\begin{small}
	\begin{align} \label{eq:noisepower1}
		\frac{1}{T+1} \mathcal{A} =&\frac{1}{T+1} \sum_{t = 1}^{T+1} \sum_{i = 1}^{N} \frac{d_{\Xc}C(Q,\epsilon)   }{2\sqrt{Q C(Q,\epsilon)     p_i} \sqrt{t}} \notag \\
		\overset{(a)}{\leq}& \frac{1}{T+1} \sum_{i = 1}^{N} \frac{d_{\Xc} C(Q,\epsilon)     }{2\sqrt{Q C(Q,\epsilon)   p_i} }2 \sqrt{T+1} \notag \\
		=&\sum_{i=1}^{N}\frac{d_{\Xc}\sqrt{C(Q,\epsilon)     }}{\sqrt{(T+1)Qp_i}},
	\end{align}
\end{small}where $(a)$ follows because
\begin{small}
	\begin{equation}
		\begin{aligned}\label{eqn:sqrt}
			\sum_{t=1}^{T+1} \frac{1}{\sqrt{t}}  \leq \int_{t=1}^{T+1} \frac{1}{\sqrt{t}} dt   \leq 2 \sqrt{T+1}.
		\end{aligned}
	\end{equation}
\end{small}
Thus, by plugging \eqref{eq:noisepower1} into \eqref{eqn:H_bounded_total_2222}, we have
\begin{small}
	\begin{align}\label{eqn:H_bounded_total_final_1_A}
		& \sum_{i =1}^{N}   \E \Big[  F_{i}( \bar{\xb}_{i}^{T+1})-F_{i}(\xb_{0}^{\star} )] + \beta \sum_{i =1}^{N} \E[\|\bar{\xb}_{i}^{T+1}-\bar{\xb}_{0}^{T+1}\| \Big]\notag\\
		\leq &  \frac{1}{T+1}  \sum_{i=1}^{N}  \sup_{\lambdab_{i} \in \Bc_{\lambda}}  \E \Big[J_{i} (\xb_{i}^{0}, \xb_{0}^{0}, \lambdab_{i})  - J_{i} (\xb_i^{T+1}, \xb_0^{T+1}, \lambdab_{i}) \Big]  \notag \\
		& +  \frac{1}{2(T+1)Q}  \sum_{i=1}^{N} \Big(    \frac{\rho  d_{\mathcal{X}}^{2} (Q+1) }{p_{i}} +     \frac{Q}{p_{i}} \frac{d_{\lambda}^{2}}{\rho}      +  \frac{4 G_{R}}{p_i} \Big) \notag \\
		&+ \frac{1}{2(T+1)Q}\sum_{i=1}^{N}\frac{\gamma_i^{T+1}d_{\Xc}^2}{p_i} +   \sum_{i=1}^{N} \frac{ d_{\mathcal{X}} \sqrt{ C(Q,\epsilon)    } }{  \sqrt{ (T+1) Q p_i}}   \notag \\
		= &  \frac{1}{T+1}  \sum_{i=1}^{N}  \sup_{\lambdab_{i} \in \mathcal{B}_{\bf \lambda}}  \big( J_{i} (\xb_{i}^{0}, \xb_{0}^{0}, \lambdab_{i})  - J_{i} (\xb_i^{T+1}, \xb_0^{T+1}, \lambdab_{i}) \big)  \notag \\
		& +  \frac{1}{2(T+1)Q}  \sum_{i=1}^{N} \Big(  \frac{\rho  d_{\mathcal{X}}^{2} (Q+1) }{p_{i}} +   \frac{Q}{p_{i}} \frac{d_{\lambda}^{2}}{\rho}      +  \frac{4 G_{R}}{p_i} \Big) \notag \\
		&+  \sum_{i=1}^{N} \frac{ 2d_{\mathcal{X}} \sqrt{ C(Q,\epsilon)    } }{  \sqrt{ (T+1) Q p_i}}.
	\end{align}
\end{small}
Note that the term
\begin{small}
	\begin{align}
		\sum_{i=1}^{N}  \sup_{\lambdab_{i} \in \mathcal{B}_{\bf \lambda}}    \E \big[J_{i} (\xb_{i}^{0}, \xb_{0}^{0}, \lambdab_{i})   - J_{i} (\xb_i^{T+1}, \xb_0^{T+1}, \lambdab_{i}) \big]
	\end{align}
\end{small}is   bounded thanks to  Assumption \ref{Ass: Assumption3}.  Thus, the RHS of \eqref{eqn:H_bounded_total_final_1_A} is also bounded.  When $Q=1$, we have the same convergence results by replacing $C(Q,\epsilon)    $ as
\begin{small}
\begin{equation}
\begin{aligned}
C(Q,\epsilon)    = G^{2}  + 2d_{\lambda}^{2} +  \frac{2\phi^{2}}{b}   + \frac{ 16 \rho d G^{2} \ln (1.25/\delta)}{  \epsilon^{2}}.
\end{aligned}
\end{equation}
\end{small}
By replacing $T+1$  in \eqref{eqn:H_bounded_total_final_1_A} as $T$,   Theorem \ref{Thm:dp theorem2} is proved. $\hfill\blacksquare$
\end{appendices}

\bibliographystyle{IEEEtran}
\bibliography{refs20,refs10}
\ifCLASSOPTIONcaptionsoff
  \newpage
\fi

\newpage
\setcounter{section}{0}

{\setcounter{equation}{0}
	\renewcommand{\theequation}{S.\arabic{equation}}

\begin{onecolumn}
\begin{center}
\LARGE
\bf Supplementary Materials: Proofs
\end{center}

\section{Proof of Lemma \ref{lem_PSI_supermum}}\label{sec: proof_of_lem_PSI_supermum}
By the definition of $\Psi(\wbb_{i}, \wt \wbb_i)$, we have
\begin{align}
		\sum_{i=1}^{N} \Psi( \bar{\wbb}_i^{T+1}, \wt \wbb_i) =& \sum_{i=1}^{N} \Big( F_{i} (\bar{\xb}_i^{T+1})-F_{i}(\xb_{i}^{\star}) +  \big\langle \bar{\wbb}_{i}^{T+1} - \wt \wbb_i, H(\bar{\wbb}_{i}^{T+1}) \big\rangle \Big) \notag \\
		\overset{(a)}{=}&  \sum_{i=1}^{N}  \Big( {F_{i}(\bar{\xb}_i^{T+1}) - F_{i}(\xb_{i}^{\star})} +  \big\langle -  \lambdab_{i}, \bar \xb_i^{T+1} - \bar \xb_{0}^{T+1} ]\big\rangle \Big) \label{lem: psi_supremum1},
\end{align}where $(a)$ follows because
\begin{align}
		\big\langle \bar{\wbb}_{i}^{T+1} - \wt \wbb_i, H(\wbb_{i}^{T+1}) \big\rangle =&	\big\langle - \bar \lambdab_i^{T+1}, \bar \xb_{i}^{T+1} - \xb_i^{\star} \big\rangle  + \big\langle \bar \lambdab_{i}^{T+1} - \lambdab_{i},  \bar \xb_i^{T+1} - \bar \xb_0  \big\rangle  +  \big\langle \bar \lambdab_{i}^{T+1},  \bar \xb_0^{T+1} - \xb_0^\star \big\rangle  \notag \\
	\overset{(b)}{=} & \big\langle -\lambdab_{i}, \bar \xb_i^{T+1} - \bar \xb_{0}^{T+1} \big\rangle, \label{lem: psi_supremum2}
\end{align}and $(b)$ holds because the constraint of $\xb_i^{\star} = \xb_0^{\star} $.  Then, by taking supremum with respect to $\lambdab_{i}$  over the two sides of \eqref{lem: psi_supremum1} yields
\begin{align}
		\sup_{\lambdab_{i} \in \Bc_{\lambda} }\sum_{i=1}^{N} \Psi( \bar \wbb_i^{T+1}, \wt \wbb_i) = & \sup_{\lambdab_{i} \in \Bc_{\lambda} } \sum_{i=1}^{N}  \Big( F_{i}\big(\bar \xb_i^{T+1}) - F_{i}(\xb_i^{\star} \big) +   \big\langle  - \lambdab_i,  \bar \xb_i^{T+1} - \bar \xb_0^{T+1}  \big\rangle \Big)\notag\\
		=&  \sum_{i=1}^{N} \Big( {F_{i}(\bar \xb_i^{T+1}) - F_{i}(\xb_i^{\star})} + \beta  \| \bar \xb_i^{T+1}- \bar \xb_0^{T+1} \|  \Big).\label{lem: psi_supremum3}
\end{align}where \eqref{lem: psi_supremum3} follows from the Assumption \ref{Ass: Assumption3}.  $\hfill\blacksquare$

\section{Proof of Lemma \ref{lem: J_connect}}\label{sec: proof_of_Lemma2}
By the definition of $J(\cdot)$ in \eqref{eqn: def_J}, we have
\begin{align}
	\E \Big[\sum_{i = 1}^{N}J_{i} (\xb_i^t, \xb_0^t, \lambdab_{i})\Big]  = &\E\Big[ \sum_{i=1}^{N} p_{i}  J_{i}(\hat \xb_i^t, \hat \xb_0^t, \lambdab_{i})  +  \sum_{i=1}^{N} (1-p_{i})  J_{i}(\xb_i^{t-1},\xb_0^{t-1}, \lambdab_{i})\Big]  \notag\\
	=& \E\Big[\sum_{i=1}^{N}   J_{i}(\xb_i^{t-1},\xb_0^{t-1},\lambdab_{i} )  +  \sum_{i=1}^{N} ( F_{i}(\hat \xb_i^t) -\langle \lambdab_{i},  \hat \xb_i^t - \hat \xb_0^t\rangle) -  \sum_{i=1}^{N} (F_{i}(\xb_i^{t-1}) - \langle \lambdab_{i},   \xb_i^{t-1} - \xb_0^{t-1} \rangle) \Big] \notag\\
	\overset{(a)}{=} & \E\Big[\sum_{i=1}^{N} J_{i}(\xb_i^{t-1},\xb_0^{t-1}, \lambdab_{i})\Big]   +  \E\Big[\sum_{i=1}^{N} \big(F_{i}(\hat \xb_i^t) - F_{i}(\xb_i^{\star}) +  \langle \hat \wbb_i^t- \wt \wbb_i,  H(\hat \wbb_i^t) \rangle \big) \Big]\notag\\
	& -  \E\Big[\sum_{i=1}^{N} \big( F_{i}(\xb_i^{t-1}) - F_{i}(\xb_i^{\star}) + \langle \wbb_i^{t-1}- \wt \wbb_i,  H(\wbb_i^{t-1}) \rangle \big) \Big] \notag\\
	\overset{(b)}{=} & \E\Big[\sum_{i=1}^{N}  \big( J_{i}(\xb_i^{t-1},\xb_0^{t-1}, \lambdab_{i}) + \Psi(\hat \wbb_i^t, \wt \wbb_i) -  \Psi(\wbb_i^{t-1}, \wt \wbb_i) \big)\Big]\label{lem: J_connect3},
\end{align}where $(a)$  follows because of \eqref{lem: psi_supremum2}; $(b)$  holds owing to the definition of $\Psi(\cdot)$ in \eqref{eqn: def_psi}. Then, we rearrange terms in the two sides of \eqref{lem: J_connect3} to obtain
\begin{align}
	\sum_{i = 1}^{N}\E \big[\Psi(\wbb_i^{t-1}, \wt \wbb_i) \big]	\leq &\sum_{i=1}^{N}  \E \big[ J_{i}(\xb_i^{t-1},\xb_0^{t-1}, \lambdab_{i}) - J_{i} (\xb_i^t, \xb_0^t, \lambdab_{i}) \big] +\sum_{ i = 1}^{N}\E\big[\Psi(\hat \wbb_i^t, \wt \wbb_i) \big].
\end{align}$\hfill\blacksquare$

\vspace{-0.5cm}
\section{Proof of Lemma \ref{lem: psi_virtual_obtain}}\label{sec: proof_of_psi_virtual_obtain}
By the definition of $\Psi(\cdot)$ and $\hat \wbb_i^t$, we have
\begin{equation}
\begin{aligned}
	 \sum_{i=1}^{N} \Psi(\hat \wbb_i^t, \wt \wbb_i)	\overset{(a)}{=} &   \sum_{i=1}^{N} \Big( {F_{i}(\hat \xb_i^t) - F_{i}(\xb_i^{\star})} +  \langle - \hat \lambdab_{i}^{t},  \hat \xb_i^t - \xb_i^{\star} \rangle   + \langle  \hat \lambdab_{i}^{t}- \lambdab_{i},  \hat \xb_i^t- \hat \xb_0^t \rangle  + \langle \hat  \lambdab_{i}^{t},  \hat \xb_0^t- \xb_0^{\star} \rangle \Big)   \\
	\overset{(b)}{\leq}  &    \sum_{i=1}^{N}\Big( \frac{1}{Q}  \sum_{r=0}^{Q-1} (F_{i}(\hat \xb_i^{t, r}) - F_{i}(\xb_i^{\star}) ) +  \langle -\hat \lambdab_{i}^{t}, \hat \xb_i^{t, r}- \xb_i^{\star}\rangle\Big) +  \sum_{i=1}^{N}  \big\langle  \hat \lambdab_{i}^{t}- \lambdab_{i},  \hat \xb_i^{t}- \hat \xb_0^{t} \big\rangle   \\
&~~+   \sum_{i=1}^{N} \big\langle  \hat \lambdab_{i}^{t}, \hat \xb_0^t - \xb_0^{\star} \big\rangle,\label{lem: psi_virtual_bound2}
\end{aligned}
\end{equation}where $(a)$ follows from the definition  $H(\wbb_i^t)$ in \eqref{eqn:def_H} by replacing $\wbb_i^t$ with $\hat \wbb_i^t$; $(b)$ holds due to the convexity of $F_i$ and the definition of $\hat \xb_i^t$ in \eqref{eqn: def_virtual_xi}. Then, we proceed to bound the three terms in the RHS of \eqref{lem: psi_virtual_bound2}. First of all, we have the following lemma which is proved in Appendix \ref{sec: proof_of_B1}.
\begin{Lemma} \label{lemma:B1}
For any $t$ and $0 \leq r \leq Q-1$, it holds that
\begin{align}
&\frac{1}{Q}  \sum_{r=0}^{Q-1}(F_{i}\Big(\hat \xb_i^{t, r}) - F_{i}(\xb_{i}^{\star}) +  \langle - \hat\lambdab_{i}^{t}, \hat \xb_i^{t, r}- \xb_i^{\star}\rangle\Big) \notag \\
\leq & \frac{\gamma_i^t + \rho}{2Q} \sum_{r=0}^{Q-1} \Big(\| \hat \xb_{i}^{t,r} -  \xb_{i}^{\star} \|^{2} -  \| \hat \xb_{i}^{t,r+1}- \xb_{i}^{\star} \|^{2} \Big)   + \frac{1}{Q}  \sum_{r=0}^{Q-1} \Big(    R_{i}(\hat \xb_{i}^{t, r}) - R_{i}(\hat \xb_{i}^{t, r+1})   +  \langle \hat{\boldsymbol\nu}_{i}^{t,r+1}, \xb_{i}^{\star} -  \hat \xb_{i}^{t, r}\rangle \Big)   \notag\\
		&+\frac{1}{Q}  \sum_{r=0}^{Q-1}  \frac{ G^{2}  + 2\|\hat{\nub}_{i}^{t,r+1}\|^{2} + 2d_{\lambda}^{2} }{ \gamma_i^t}, \label{lem: psi_virtual_bound3}
\end{align}
\end{Lemma}where $\hat{ \boldsymbol \nu}_{i}^{t,r+1} = \nabla f_{i}(\hat \xb_i^{t, r}; B_{i}^{t,r+1}) - \nabla f_{i}(\hat \xb_i^{t, r})$. Next, using the definition of $\hat \lambdab_{i}^t$ in \eqref{eqn: def_virtual_dual}, we get
\begin{align}
	\big\langle  \hat \lambdab_{i}^{t} - \lambdab_{i},  \hat \xb_i^t  - \hat \xb_0^t \big\rangle = & \frac{1}{\rho} \big\langle \hat \lambdab_{i}^{t}- \lambdab_{i},  \lambdab_{i}^{t-1}-\hat \lambdab_{i}^{t} \big\rangle  \notag \\
	= & \frac{1}{2\rho} \Big(\|\lambdab_{i} - \lambdab_{i}^{t-1}\|^{2} - \|\hat \lambdab_{i}^{t} -  \lambdab_{i}^{t-1} \|^{2}- \|\lambdab_{i} - \hat \lambdab_{i}^{t}\|^{2}\Big). \label{lem: psi_virtual_bound5}
\end{align}
In addition, the last term in the RHS of \eqref{lem: psi_virtual_bound2} can be bounded by the following lemma.

\begin{Lemma}\label{lemma:E1} If $ \rho \geq \ \sqrt{\gamma_i^{t}}, \forall t$,   it holds that
\begin{align}
			 \sum_{i=1}^{N}   \big\langle \hat \lambdab_{i}^{t}, \hat \xb_0^t - \xb_0^{\star} \big\rangle \leq  &  \frac{\rho}{2} \sum_{i=1}^{N} \big(\|\xb_0^{\star} -  \xb_i^{t-1}\|^{2}  -   \|\xb_0^{\star} - \hat \xb_i^t\|^{2}\big)  + \rho \sum_{i=1}^{N} \big\langle \boldsymbol{\xi}_{i}^{t-1}, \xb_i^{t-1} - \xb_0^{\star} \big\rangle  \notag \\
			& +  \sum_{i=1}^{N} \frac{\rho}{2\gamma_i^t}\| ( \rho + \gamma_i^{t-1}) \xib_i^{t-1}\|^2 + \frac{1}{2\rho} \sum_{i=1}^{N}\| \hat \lambdab_{i}^{t}- \hat \lambdab_{i}^{t-1}\|^{2}. \label{lem: psi_virtual_bound6}
\end{align}
\end{Lemma} Substituting the intermediate results \eqref{lem: psi_virtual_bound3}, \eqref{lem: psi_virtual_bound5} and \eqref{lem: psi_virtual_bound6} into \eqref{lem: psi_virtual_bound2}, and then taking expectation over the two sides of \eqref{lem: psi_virtual_bound2} gives rise to
\begin{align}
\E\Big[\sum_{i=1}^{N} \Psi(\hat \wbb_i^t, \wt \wbb_i)\Big]	\leq &   \E\bigg[\frac{1}{Q} \sum_{i=1}^{N}  \sum_{r=0}^{Q-1}  \Big(  \big(\gamma_i^t)^{-1} ( G^{2}  +  2\| \hat \nub_{i}^{t, r+1}\|^{2} + 2d_{\lambda}^{2}  +   \frac{\rho}{2} \| ( \rho + \gamma_i^{t-1}) \xib_i^{t-1} \|^{2} \big)   \notag \\
&~+  \frac{\gamma_i^t+\rho}{2} \big(\| \hat \xb_i^{t, r} -   \xb_i^{\star} \|^{2} -  \|\hat \xb_i^{t, r+1}-  \xb_i^{\star} \|^{2} \big) +  \frac{\rho}{2} \big(\| \xb_0^{\star} - \xb_{i}^{t-1}\|^{2}  -   \|\xb_0^{\star} - \hat{\xb}_{i}^{t}\|^{2} \big)   \notag  \\
&~~~~+ \frac{1}{2\rho} \big(\| \lambdab_{i} - \lambdab_{i}^{t-1}\|^{2}  - \| \lambdab_{i} - \hat \lambdab_{i}^t \|^{2} \big)  +  \big(R_{i}(\hat \xb_i^{t, r}) -    R_{i}(\hat \xb_i^{t, r+1}) \big)   \notag   \\
&~~~~+ \rho \big\langle \xib_i^{t-1}, \xb_{i}^{t-1} - \xb_0^{\star} \big\rangle + \big\langle \hat \nub_{i}^{t, r+1}, \xb_i^{\star} -  \hat{\xb}_{i}^{t, r} \big\rangle \Big)\bigg] \notag \\
\overset{(c)}{\leq}&\E\bigg[\frac{1}{Q} \sum_{i=1}^{N}  \sum_{r=0}^{Q-1}   \Big(  \big(\gamma_i^t)^{-1} ( G^{2}  +  \frac{2\phi^2}{b} + 2d_{\lambda}^{2}  +   \frac{\rho}{2} \| ( \rho + \gamma_i^{t-1}) \xi_i^{t-1} \|^{2} \big)   \notag \\
&~~~~ +  \frac{\gamma_i^t+\rho}{2} \big(\| \hat \xb_i^{t, r} -   \xb_i^{\star} \|^{2} -  \|\hat \xb_i^{t, r+1}-  \xb_i^{\star} \|^{2} \big) +  \frac{\rho}{2} \big(\| \xb_0^{\star} - \xb_{i}^{t-1}\|^{2}  -   \|\xb_0^{\star} - \hat{\xb}_{i}^{t}\|^{2} \big) \notag    \\
&~~~~ + \frac{1}{2\rho} \big(\| \lambdab_{i} - \lambdab_{i}^{t-1}\|^{2}  - \| \lambdab_{i} - \hat \lambdab_{i}^t \|^{2} \big)  +  \big(R_{i}(\hat \xb_i^{t, r}) -    R_{i}(\hat \xb_i^{t, r+1}) \big) \Big)\bigg] \label{lem: psi_virtual_bound8},
\end{align}where $(c)$ follows by Assumption \ref{Ass: Assumption2} and the fact that $\boldsymbol{\xi}_{i}^{t-1}$ is sampled from a Gaussian distribution with zero mean, i.e.,
\begin{align}
&\E\big[\| \hat{ \nub}_{i}^{t, r+1}\|^{2}\big] \leq \frac{\phi^2}{b},~
\E\big[\langle \hat{ \nub}_{i}^{t, r+1}, \xb_i^{\star} - \hat \xb_i^{t, r} \rangle\big] = 0, \\
&\E\big[\langle \xib_i^{t-1}, \xb_{i}^{t-1} - \xb_0^{\star}\rangle\big] = 0.
\end{align}
Then, by summing the inequality \eqref{lem: psi_virtual_bound8} from $t = 1$ to $T+1$, and then divide it by $T+1$ to obtain
\begin{equation}
\begin{aligned}
	&\frac{1}{T+1}\sum_{t = 1}^{T}\sum_{i=1}^{N}\E\big[ \Psi(\hat \wbb_i^t, \wt \wbb_i)\big]   \\
	\overset{(d)}{\leq} & \frac{1}{T+1}   \sum_{t = 1}^{T+1}\sum_{i = 1}^{N} \frac{1}{\gamma_i^{t}} \Big( G^{2}  +  \frac{2\phi^2}{b} + 2d_{\lambda}^{2}  + \frac{\rho}{2}\E \big[\| ( \rho + \gamma_i^{t-1}) \xib_i^{t-1} \|^{2} \big]  \Big) 	  \\
 & + \frac{1}{(T+1)Q}\sum_{t = 1}^{T+1}\sum_{i = 1}^{N} \frac{ \gamma_i^{T+1} + \rho}{2} \E\big[ \| \xb_i^{t-1,Q} -   \xb_i^{\star} \|^{2} -  \|\hat \xb_i^{t, Q}-  \xb_i^{\star} \|^{2} \big] +  \frac{\rho}{2}\frac{1}{T+1}\sum_{t = 1}^{T+1} \sum_{i = 1}^{N}\E \big[\| \xb_0^{\star} - \xb_{i}^{t-1}\|^{2}  -   	\|\xb_0^{\star} - \hat{\xb}_{i}^{t}\|^{2} \big]      \\
	& + \frac{1}{2\rho}\frac{1}{T+1}\sum_{t = 1}^{T+1}\sum_{i = 1}^{N}\E \big[ \|\lambdab_{i} - \lambdab_{i}^{t-1}\|^{2}  - \|\lambdab_{i} - \hat \lambdab_{i}^t \|^{2} \big] + \frac{1}{(T+1)Q}\sum_{t = 1}^{T+1}\sum_{i = 1}^{N} \E\big[(R_{i}( \xb_i^{t-1,Q}) -  R_{i}(\hat \xb_i^{t, Q}))\big] 	  \\
\overset{(e)}{\leq} & \frac{1}{T+1}\sum_{t = 1}^{T+1} \sum_{i = 1}^{N} \frac{1}{\gamma_i^{t}} \Big( G^{2}  +  \frac{2\phi^2}{b} + 2d_{\lambda}^{2} ) + \frac{\rho}{2} \E\big[ \| ( \rho + \gamma_i^{t-1})\xi_i^{t-1} \|^{2} \big] \Big)  	  \\
& + \frac{1}{(T+1)Q} \sum_{i = 1}^{N} \frac{ \gamma_i^{T+1} + \rho }{2p_i} \E \big[\|  \xb_i^{0,Q}  -   \xb_i^{\star} \|^{2}  - \|  \xb_i^{T+1,Q} -   \xb_i^{\star} \|^{2} \big] +  \frac{\rho}{2} \frac{1}{T+1} \sum_{i = 1}^{N}\frac{1}{p_i}\E\big[\|\xb_i^{0} - \xb_0^{\star}\|^2 - \|\xb_i^{T+1} - \xb_0^{\star}\|^2 \big]      \\
& + \frac{1}{2\rho}\frac{1}{T+1} \sum_{i = 1}^{N}\frac{1}{p_i} \E\big[\|\lambdab_i^{0} - \lambdab_{i}\|^2 - \|\lambdab_i^{T+1} - \lambdab_i^{\star}\|^2\big]  + \frac{1}{(T+1)Q}\sum_{i = 1}^{N} \frac{1}{p_i} \E [R_{i}(  \xb_i^{0,Q}) -  R_{i}( \xb_i^{T+1, Q})],
\end{aligned}
\end{equation}where $(d)$ follows because $\hat \xb_i^{t,0} = \xb_i^{t-1,Q}$, $\gamma_i^t \leq \gamma_i^{T+1}$,  and  $(e)$ holds because

\begin{align}
\sum_{i=1}^{N}  \frac{1}{p_{i}}\E \big[ \|\xb_i^t -  \xb_0^{\star} \|^{2} \big] \overset{(f)}{=} &  \sum_{i=1}^{N}\E \big[\|\hat \xb_i^t -    \xb_0^{\star}\|^{2} ]  +  \sum_{i=1}^{N} (1- p_{i}) \frac{1}{p_{i}}\E [\|  \xb_i^{t-1} -   \xb_0^{\star} \|^{2} \big] \notag\\
		=& \sum_{i=1}^{N}   \frac{1}{p_{i}} \E \big[\|  \xb_i^{t-1} -    \xb_0^{\star} \|^{2} \big]  + \sum_{i=1}^{N}  \E \big[\| \hat \xb_i^t -    \xb_0^{\star} \|^{2}- \|  \xb_i^{t-1} -    \xb_0^{\star}\|^{2} \big], \label{lem: partial_var_b1}
\end{align}where $(f)$ holds since $\xb_i^t = \hat \xb_{i}^t$ with probability $p_i$, and   $\xb_i^t = \xb_{i}^{t-1}$ with a probability $1-p_i$. \eqref{lem: partial_var_b1} further implies
\begin{align}
	\sum_{t = 1}^{T+1}\sum_{i = 1}^{N}\E\big[\|\xb_i^{t-1} - \xb_0^{\star}\|^2 - \|\hat \xb_i^t - \xb_0^{\star}\|^2 \big] =&  \sum_{t = 1}^{T+1}\sum_{i = 1}^{N}\frac{1}{p_i}\E \big[\|\xb_i^{t-1} - \xb_0^{\star}\|^2 - \|\xb_i^t - \xb_0^{\star}\|^2 \big] \notag \\
		=&  \sum_{i = 1}^{N}\frac{1}{p_i}\E \big[\|\xb_i^{0} - \xb_0^{\star}\|^2 - \|\xb_i^{T+1}- \xb_0^{\star}\|^2 \big]. \label{lem: partial_var_b11}		
\end{align}
Similarly, $ \E\big[\|\xb_i^{t-1,Q} - \xb_0^{\star}\|^2 - \|\hat \xb_i^{t,Q} - \xb_0^{\star}\|^2\big]$, $ \E\big[\|\lambdab_i^{t-1} - \lambdab_{i}\|^2 - \|\hat \lambdab_i^t - \lambdab_{i}\|^2\big]$ and  $ \E\big[R_i(  \xb_i^{t-1,Q}) - R_i(\hat \xb_i^{t,Q})\big]$ can be bounded in the same manner. $\hfill\blacksquare$

\section{Proof of Lemma \ref{lemma:B1}}\label{sec: proof_of_B1}
By the definition of $\hat \wbb_i^t $,  we have,
\begin{align} \label{eqn: objective_sub_11}
&\frac{1}{Q}  \sum_{r=0}^{Q-1} \Big( {F_{i}(\hat \xb_{i}^{t, r}) - F_{i}(\xb_{i}^{\star})} +  \big\langle - \hat\lambdab_{i}^{t}, \hat \xb_{i}^{t, r}- \xb_{i}^{\star} \big\rangle \Big) \notag \\
=& \frac{1}{Q}  \sum_{r=0}^{Q-1} \Big( f_{i} (\hat \xb_{i}^{t, r}) - f_{i}(\xb_{i}^{\star})   + \big\langle - \hat \lambdab_{i}^{t}, \hat \xb_{i}^{t, r}- \xb_{i}^{\star} \big\rangle   + R_{i}(\hat \xb_{i}^{t, r}) - R_{i}(\xb_{i}^{\star}) \Big) \notag \\
		\overset{(a)}{\leq} & \frac{1}{Q}  \sum_{r=0}^{Q-1} \Big( \big\langle \nabla f_{i}(\hat \xb_{i}^{t, r}),  \hat \xb_{i}^{t, r} -  \xb_{i}^{\star} \big\rangle +  \big\langle - \hat\lambdab_{i}^{t}, \hat \xb_{i}^{t, r}- \xb_{i}^{\star} \big\rangle  +  \big\langle \zeta_{i}^{t,r+1},  \hat \xb_{i}^{t, r+1} - \xb_{i}^{\star} \big\rangle  +  R_{i}(\hat \xb_{i}^{t, r}) - R_{i}(\hat \xb_{i}^{t, r+1})\Big) \notag \\
		= &    \underbrace{  \frac{1}{Q}  \sum_{r=0}^{Q-1} \big\langle\nabla f_{i}\left(\hat \xb_{i}^{t, r},B_{i}^{t,r+1}\right) +  \zeta_{i}^{t,r+1}, \hat \xb_{i}^{t, r+1}-\xb_{i}^{\star} \big\rangle }_{A_{1}}   + \underbrace{ \frac{1}{Q}  \sum_{r=0}^{Q-1}\big\langle\nabla f_{i}\left(\hat \xb_{i}^{t, r}\right), \xb_{i}^{t, r}- \hat \xb_{i}^{t, r+1}\big\rangle}_{A_{2}} \notag \\
& +   \underbrace { \frac{1}{Q}  \sum_{r=0}^{Q-1} \big\langle \hat {\boldsymbol\nu_{i}}^{t,r+1}, \xb_{i}^{\star} - \hat \xb_{i}^{t, r+1}\big\rangle}_{A_{3}}  + \frac{1}{Q}  \sum_{r=0}^{Q-1} \Big(  R_{i}(\hat \xb_{i}^{t, r}) - R_{i}(\hat \xb_{i}^{t, r+1})  +  \big\langle - \hat \lambdab_{i}^{t}, \hat \xb_{i}^{t, r}- \xb_{i}^{\star} \big\rangle\Big),
\end{align}where $ \zeta_{i}^{t,r+1} \in \partial R_{i} (\hat{ \xb_{i}}^{t,r+1} )$ is a subgradient of $R_{i}( \cdot )$ at $\hat \xb_{i}^{t,r+1}$; $\hat {\boldsymbol\nu_{i}}^{t,r+1} \triangleq  \nabla f_{i}\big(\hat \xb_{i}^{t, r},B_{i}^{t,r+1}\big) - \nabla f_{i}\big(\hat \xb_{i}^{t, r}\big)$; the inequality $(a)$ follows by  the convexity of $f(\cdot)$ and $R(\cdot)$, i.e.,

\begin{align} \label{eqn:R}
		f_{i} (\hat \xb_{i}^{t, r}) - f_{i}(\xb_{i}^{\star}) \leq &	\big\langle \nabla f_{i}(\hat \xb_{i}^{t, r}),  \hat \xb_{i}^{t, r} -  \xb_{i}^{\star} \big\rangle ,\\
		 R_{i}(\hat \xb_{i}^{t, r}) - R_{i}(\xb_{i}^{\star})=& R_{i}(\hat \xb_{i}^{t, r}) - R_{i}(\hat \xb_{i}^{t, r+1}) +   R_{i}(\hat \xb_{i}^{t, r+1}) - R_{i}(\xb_{i}^{\star})\notag\\
		\leq & R_{i}(\hat \xb_{i}^{t, r}) - R_{i}(\hat \xb_{i}^{t, r+1})   + \big\langle \zeta_{i}^{t,r+1},  \hat \xb_{i}^{t, r+1} - \xb_{i}^{\star} \big\rangle.
\end{align}
We proceed to bound the terms $A_{1}$,  $A_{2}$, and $A_{3}$ respectively.  For $A_{1}$, it holds that
\begin{align}\label{eqn:B1}
		& \frac{1}{Q}  \sum_{r=0}^{Q-1} \big\langle \nabla f_{i}(\hat \xb_{i}^{t, r},B_{i}^{t,r+1}) +  \zeta_{i}^{t,r+1}, \hat \xb_{i}^{t, r+1}- \xb_{i}^{\star} \big\rangle \notag \\
		\overset{(b)}{=} &\frac{1}{Q}  \sum_{r=0}^{Q-1}\gamma^{t} \big\langle \xb_{i}^{t, r}- \xb_{i}^{t, r+1}, \xb_{i}^{t, r+1}- \xb_{i}^{\star} \big\rangle +\frac{1}{Q}  \sum_{r=0}^{Q-1}\big\langle\lambdab_{i}^{t-1}- \rho (\hat \xb_{i}^{t, r+1}- \xb_{0}^{t} ), \hat \xb_{i}^{t, r+1}- \xb_{i}^{\star} \big\rangle \notag \\
		= &\frac{1}{Q}  \sum_{r=0}^{Q-1}\frac{\gamma^{t}}{2} \big(\| \hat \xb_{i}^{t,r} -  \xb_{i}^{\star} \|^{2} -   \|\hat \xb_{i}^{t,r+1}- \hat\xb_{i}^{t,r}\|^{2}
		-\| \hat \xb_{i}^{t,r+1} -  \xb_{i}^{\star} \|^{2} \big) \notag \\
		&~~~+\frac{1}{Q} \sum_{r=0}^{Q-1} \big\langle \lambdab_{i}^{t-1}-\rho  (\hat \xb_{i}^{t}- \hat \xb_{0}^{t} )+ \rho (\hat \xb_i^t- \hat\xb_{0}^{t}) -    \rho (\hat \xb_{i}^{t, r+1}- \hat \xb_{0}^{t}), \hat \xb_{i}^{t, r+1}- \xb_{i}^{\star} \big\rangle \notag \\
		\overset{(c)}{=} &\frac{1}{Q}  \sum_{r=0}^{Q-1}\frac{\gamma^{t}}{2} \big(\| \hat \xb_{i}^{t,r} -  \xb_{i}^{\star} \|^{2} -   \|\hat \xb_{i}^{t,r+1}- \hat\xb_{i}^{t,r}\|^{2}
		-\| \hat \xb_{i}^{t,r+1} -  \xb_{i}^{\star} \|^{2} \big) + \frac{\rho}{Q} \sum_{r=0}^{Q-1}  \big\langle \hat \xb_{i}^{t}- \hat \xb_{i}^{t, r+1}, \hat \xb_{i}^{t, r+1}- \xb_{i}^{\star} \big\rangle \notag \\
		&~~~+ \frac{1}{Q} \sum_{r=0}^{Q-1} \big\langle \hat \lambdab_{i}^{t}, \hat \xb_{i}^{t,r+1}- \xb_{i}^{\star} \big\rangle \notag \\
		\overset{(d)}{\leq}& \frac{1}{Q}  \sum_{r=0}^{Q-1}\frac{\gamma^{t}}{2} \big(\| \hat \xb_{i}^{t,r} -  \xb_{i}^{\star} \|^{2} -   \|\hat \xb_{i}^{t,r+1}- \hat\xb_{i}^{t,r}\|^{2}
		-\| \hat \xb_{i}^{t,r+1} -  \xb_{i}^{\star} \|^{2} \big) + \frac{\rho}{2Q}\sum_{r = 0}^{Q-1} \big(\| \hat \xb_{i}^{t,r} -  \xb_{i}^{\star} \big\|^{2} -  \| \hat \xb_{i}^{t,r+1}- \xb_{i}^{\star} \|^{2} \big) \notag \\
		&~~~+ \frac{1}{Q} \sum_{r=0}^{Q-1} \big\langle \hat \lambdab_{i}^{t}, \hat \xb_{i}^{t,r+1}- \xb_{i}^{\star} \big\rangle,
\end{align}where  $(b)$ follows by the optimality condition of  (\ref{eqn: local update_primal_1}),  i.e.,
\begin{align} \label{eqn:C1}
		 \nabla f_{i} ( \hat \xb_{i}^{t,r}, B_{i}^{t,r+1}) +  \zeta_{i}^{t,r+1}- \lambdab_{i}^{t-1}  + \rho(\hat \xb_{i}^{t,r+1} - \xb_{0}^{t}) = & \gamma^{t} \big(\hat \xb_{i}^{t,r} - \hat \xb_{i}^{t,r+1}\big),
\end{align}$(c)$ follows because of \eqref{eqn: admm-update2_c}, and $(d)$ holds by
\begin{align}\label{eqn:D1_1}
		\frac{1}{Q}\sum_{r=0}^{Q-1}  \big\langle \hat \xb_{i}^{t}- \hat \xb_{i}^{t, r+1}, \hat \xb_{i}^{t, r+1}   - \xb_{i}^{\star} \big\rangle  = & \frac{1}{Q^2} \sum_{r=0}^{Q-1}  \sum_{l=0}^{Q-1} \big\langle \hat \xb_{i}^{t,l}- \hat \xb_{i}^{t, r+1}, \hat \xb_{i}^{t, r+1}- \xb_{i}^{\star} \big\rangle \notag  \\
		= & \frac{1}{2Q^{2}} \sum_{r=0}^{Q-1} \sum_{l=0}^{Q-1} \big( \| \hat \xb_{i}^{t, l} - \xb_{i}^{\star} \|^{2} -\| \hat\xb_{i}^{t, r+1} - \hat\xb_{i}^{t, l}\|^{2}- \| \hat \xb_{i}^{t, r+1} - \xb_{i}^{\star}\|^{2} \big) \notag \\
		\leq & \frac{1}{2Q^{2}} \sum_{r=0}^{Q-1} \sum_{l=0}^{Q-1} \big( \| \hat \xb_{i}^{t, l} - \xb_{i}^{\star} \|^{2} - \| \hat \xb_{i}^{t, r+1} - \xb_{i}^{\star}\|^{2} \big) \notag \\
		= & \frac{1}{2}\Big( \frac{1}{Q}\sum_{l=0}^{Q -1}\| \hat \xb_{i}^{t, l} - \xb_{i}^{\star} \|^{2} -  \frac{1}{Q}\sum_{r=0}^{Q-1}\| \hat \xb_{i}^{t, r+1} - \xb_{i}^{\star}\|^{2}\Big)\notag \\
		= & \frac{1}{2Q} \sum_{r = 0}^{Q-1 }\big(\| \hat \xb_{i}^{t,r} -  \xb_{i}^{\star} \big\|^{2} -  \| \hat \xb_{i}^{t,r+1}- \xb_{i}^{\star} \|^{2} \big).
\end{align}
Next, for the term $A_{2}$, it follows that
\begin{align} \label{eqn:A_2_1}
	\frac{1}{Q}  \sum_{r=0}^{Q-1} \big\langle \nabla f_{i}(\hat \xb_{i}^{t, r}), \hat\xb_{i}^{t, r}- \hat\xb_{i}^{t, r+1} \big\rangle \overset{(e)}{\leq}  &   \frac{1}{Q}  \sum_{r=0}^{Q-1} \Big(   \frac{\| \nabla f_{i}\hat \xb_{i}^{t, r})\|^{2}}{ \gamma^{t}}  + \frac{\gamma^{t} \|\hat \xb_{i}^{t, r}- \hat \xb_{i}^{t, r+1}\|^{2}}{4} \Big) \notag \\
\overset{(f)}{\leq} &   \frac{1}{Q}  \sum_{r=0}^{Q-1} \Big( \frac{ G^{2}  }{ \gamma^{t}}  + \frac{\gamma^{t} \big\|\hat \xb_{i}^{t, r}- \hat \xb_{i}^{t, r+1}\big\|^{2}}{4} \Big),
\end{align}where $(e)$ holds thanks to the Young's inequality, i.e.,  $\langle \xb, \tilde{{\bm y}}\rangle \leq \frac{1}{2c}\|\xb\|^{2}+\frac{c}{2}\|\tilde{{\bm y}}\|^{2}$ for any $c >0$, and $(f)$ follows by \eqref{eqn:bounded_gradient}. The term $A_{3}$ can be bounded by
\begin{align}\label{eqn:A3_1}
	 \frac{1}{Q}  \sum_{r=0}^{Q-1} \big\langle \hat{ \nub}_{i}^{t,r+1}, \xb_{i}^{\star} - \hat \xb_{i}^{t, r+1} \big\rangle  = & \frac{1}{Q}  \sum_{r=0}^{Q-1} \Big(\big\langle \hat{ \nub}_{i}^{t,r+1}, \xb_{i}^{\star} - \hat \xb_{i}^{t, r} \big\rangle +   \big\langle \hat {\nub}_{i}^{t,r+1}, \hat \xb_{i}^{t, r} - \hat \xb_{i}^{t, r+1}\big\rangle \Big)\notag\\
	\overset{(g)}{\leq} &  \frac{1}{Q}  \sum_{r=0}^{Q-1} \Big(  \big\langle \hat{ \nub}_{i}^{t,r+1}, \xb_{i}^{\star} -  \hat \xb_{i}^{t, r} \big\rangle + \frac{2 \| \hat {\nub}_{i}^{t,r+1}\|^{2}}{ \gamma^{t}} + \frac{\gamma^{t} \big\| \hat \xb_{i}^{t, r}- \hat \xb_{i}^{t, r+1}\big\|^{2}}{8} \Big),
\end{align}where $(g)$ follows by the Young's inequality. Substituting the results of \eqref{eqn:B1},  \eqref{eqn:A_2_1}, and \eqref{eqn:A3_1}) into \eqref{eqn: objective_sub_11} gives rise to
\begin{align}  \label{eqn:B1_result}
		&  \frac{1}{Q}  \sum_{r=0}^{Q-1} \big( {F_{i}(\hat \xb_{i}^{t, r}) - F_{i}(\xb_{i}^{\star})} +  \langle - \hat \lambdab_{i}^{t}, \hat \xb_{i}^{t, r}- \xb_{i}^{\star} \rangle \big)\notag\\
		\leq & \frac{\rho + \gamma^t}{2Q} \sum_{r=0}^{Q-1} \frac{\gamma^{t}}{2}  \big(\| \hat \xb_{i}^{t,r} -  \xb_{i}^{\star} \|^{2} -  \| \hat \xb_{i}^{t,r+1}- \xb_{i}^{\star} \|^{2}\big) + \frac{1}{Q}  \sum_{r=0}^{Q-1} \big(   R_{i}( \hat \xb_{i}^{t, r}) - R_{i}( \hat \xb_{i}^{t, r+1}) + \langle \hat \lambdab_{i}^{t}, \hat \xb_{i}^{t, r+1}- \hat \xb_{i}^{t, r} \rangle \big) \notag\\
		&+\frac{1}{Q}  \sum_{r=0}^{Q-1} \big( \frac{G^{2} + 2\|\hat { \nub}_{i}^{t,r+1}\|^{2} }{ \gamma^{t}} -\frac{\gamma^{t} \| \hat \xb_{i}^{t, r}- \hat \xb_{i}^{t, r+1}\big\|^{2}}{8} \big) + \frac{1}{Q}  \sum_{r=0}^{Q-1}    \big\langle \hat{\nub}_{i}^{t,r+1}, \xb_{i}^{\star} -  \hat \xb_{i}^{t, r} \big\rangle  \notag\\
		\overset{(h)}{\leq} & \frac{\gamma_i^t + \rho}{2Q} \sum_{r=0}^{Q-1} \big(\| \hat \xb_{i}^{t,r} -  \xb_{i}^{\star} \|^{2} -  \| \hat \xb_{i}^{t,r+1}- \xb_{i}^{\star} \|^{2}\big)  + \frac{1}{Q}  \sum_{r=0}^{Q-1} \big(    R_{i}(\hat \xb_{i}^{t, r}) - R_{i}(\hat \xb_{i}^{t, r+1})   +  \big\langle \hat{\boldsymbol\nu}_{i}^{t,r+1}, \xb_{i}^{\star} -  \hat \xb_{i}^{t, r}\big\rangle \big)   \notag\\
		&+\frac{1}{Q}  \sum_{r=0}^{Q-1}  \frac{ G^{2}  + 2\|\hat{\nub}_{i}^{t,r+1}\|^{2} + 2d_{\lambda}^{2} }{ \gamma_i^t},
\end{align}where $(h)$ follows because of Assumption \ref{Ass: Assumption3} and
\begin{small}
	\begin{align}\label{eqn:C_2}
		\big\langle \hat \lambdab_{i}^{t}, \hat \xb_{i}^{t, r+1}- \hat \xb_{i}^{t, r} \big\rangle &\leq \frac{2 \|\hat \lambdab_{i}^{t}\|^{2}}{ \gamma^{t}} + \frac{\gamma^{t} \| \hat \xb_{i}^{t, r}- \hat \xb_{i}^{t, r+1}\|^{2}}{8} \notag \\
		& \leq \frac{2 d_{\lambda}^{2}}{ \gamma^{t}} + \frac{\gamma^{t} \| \hat \xb_{i}^{t, r}- \hat \xb_{i}^{t, r+1}\|^{2}}{8}.
	\end{align}
\end{small}$\hfill\blacksquare$
\vspace{-0.3cm}
\subsection{Proof of Lemma \ref{lemma:E1}}\label{sec: Proof_of_E1}
According to the  \eqref{eqn: admm-update2_a} and   \eqref{eqn: admm-update2_c},  we have
\begin{subequations}
\begin{align}
\hat \xb_{0}^{t} &= \frac{1}{\rho} ( \hat \lambdab_{i}^{t}-\lambdab_{i}^{t-1} ) + \hat \xb_{i}^{t}, \label{eqn:A2_handling_a}\\
\hat \xb_{0}^{t} & = \frac{1}{N} \sum_{i=1}^{N} \big( \xb_{i}^{t-1}-  \frac{\lambdab_{i}^{t-1}}{\rho}  + \xib_{i}^{t-1} \big).   \label{eqn:A2_handling_b1}
\end{align}
\end{subequations}
Summing the equalities \eqref{eqn:A2_handling_a} and \eqref{eqn:A2_handling_b1} from $i = 1$ to $N$, and then combing them together yield
\vspace{-0.2cm}
\begin{align}
\sum_{i=1}^{N}  \big(\frac{1}{\rho} ( \hat \lambdab_{i}^{t}-\lambdab_{i}^{t-1} ) + \hat \xb_{i}^{t}\big) = \sum_{i=1}^{N} \big( \xb_{i}^{t-1}-  \frac{\lambdab_{i}^{t-1}}{\rho}    + \xib_{i}^{t-1} \big),
\end{align}
which implies that
\begin{align}
\sum_{i=1}^{N}  \Big(  \frac{\hat \lambdab_{i}^{t}}{\rho}  + \hat \xb_{i}^{t} -  \xb_{i}^{t-1} - \xib_{i}^{t-1} \Big) =0. \label{eqn:key_2}
\end{align}
Using \eqref{eqn:key_2},  we have
\begin{equation}
\begin{aligned}\label{eqn:E1}
\sum_{i=1}^{N}  \big\langle \hat \lambdab_{i}^{t}, \hat \xb_{0}^{t} - \xb_{0}^{\star} \big\rangle = &   \rho \sum_{i=1}^{N}   \big\langle \xb_{i}^{t-1} -  \hat \xb_{i}^{t}, \hat \xb_{0}^{t}-\xb_{0}^{\star} \big\rangle    + \rho \sum_{i=1}^{N}  \big\langle \xib_{i}^{t-1}, \hat \xb_{0}^{t} - \xb_{0}^{\star} \big\rangle     \\
= & \frac{\rho}{2} \sum_{i=1}^{N}   ( \| \hat \xb_{0}^{t}- \hat \xb_{i}^{t}\|^{2} - \| \hat \xb_{0}^{t}- \xb_{i}^{t-1}\|^{2} +  \|\xb_{0}^{\star} - \xb_{i}^{t-1}\|^{2}  -    \|\xb_{0}^{\star} - \hat \xb_{i}^{t}\|^{2} ) +  \rho \underbrace{ \sum_{i=1}^{N}  \big\langle \boldsymbol{\xi}_{i}^{t-1}, \hat \xb_{0}^{t} - \xb_{0}^{\star} \big\rangle}_{B_1}.
\end{aligned}
\end{equation}
We proceed to bound the $B_1$ term in RHS of \eqref{eqn:E1} by
\begin{align}
\sum_{i=1}^{N}   \big\langle \xib_{i}^{t-1}, \hat \xb_{0}^{t} - \xb_{0}^{\star} \big\rangle = & \sum_{i=1}^{N}   \big\langle \xib_{i}^{t-1}, \xb_{i}^{t-1} - \xb_{0}^{\star} \big\rangle +  \sum_{i=1}^{N}  \big\langle \xib_{i}^{t-1}, \hat \xb_{0}^{t} - \xb_{i}^{t-1} \big\rangle \notag\\
\overset{(a)}{\leq} &  \sum_{i=1}^{N} \Big( \frac{1}{2 \beta^{t}} \| \hat \xb_{0}^{t} - \xb_{i}^{t-1} \|^{2}  +  \frac{\beta^{t}}{2}\|\xib_i^{t-1}\|^2 \Big) + \sum_{i=1}^{N}    \big\langle \xib_{i}^{t-1}, \xb_{i}^{t-1} - \xb_{0}^{\star} \big\rangle, \label{eqn:E3}
\end{align}where $(a)$  follows by the Young's inequality and $\beta^{t} \triangleq  ( \rho + \gamma_i^{t-1})^{2}/\gamma_i^t$. Thus, we have
\begin{align}\label{eqn:E1_sum}
\sum_{i=1}^{N}   \big\langle \hat \lambdab_{i}^{t}, \hat \xb_{0}^{t} - \xb_{0}^{\star} \big\rangle  \leq  &  \frac{\rho}{2} \sum_{i=1}^{N} \Big(\| \hat \xb_{0}^{t}- \hat \xb_{i}^{t}\|^{2} +  \|\xb_{0}^{\star} - \xb_{i}^{t-1}\|^{2}  -    \|\xb_{0}^{\star} -  \hat \xb_{i}^{t}\|^{2}  +    \beta^t\|\xib_{i}^{t-1}\|^2 \notag \\
&~~ - \Big(1- \frac{1}{\beta^t}\Big) \|  \hat \xb_{0}^{t} - \xb_{i}^{t-1} \|^{2}   + 2\langle \xib_{i}^{t-1}, \xb_{i}^{t-1} - \xb_{0}^{\star} \rangle \Big) \notag\\
\overset{(b)}{\leq}  &  \frac{\rho}{2} \sum_{i=1}^{N} \big( \| \hat \xb_{0}^{t}- \hat \xb_{i}^{t}\|^{2} +  \|\xb_{0}^{\star} - \xb_{i}^{t-1}\|^{2}  -   \|\xb_{0}^{\star} - \hat \xb_{i}^{t}\|^{2} \big) + \frac{\rho}{2} \sum_{i=1}^{N} \big(\beta^t \|\xib_i^{t-1}\|^2+ 2\langle \xib_{i}^{t-1}, \xb_{i}^{t-1} - \xb_{0}^{\star} \rangle \big) \notag \\
\overset{(c)}{=} &  \frac{\rho}{2} \sum_{i=1}^{N} \big( \|\xb_{0}^{\star} - \xb_{i}^{t-1}\|^{2}  -    \|\xb_{0}^{\star} - \hat \xb_{i}^{t}\|^{2} \big) + \rho \sum_{i=1}^{N} \big\langle \xib_{i}^{t-1}, \xb_{i}^{t-1} - \xb_{0}^{\star} \big\rangle  \notag\\
& +  \sum_{i=1}^{N} \frac{\rho\beta^t}{2}\|\xib_i^{t-1}\|^2 + \frac{1}{2\rho} \sum_{i=1}^{N}     \| \hat \lambdab_{i}^{t}- \lambdab_{i}^{t-1}\|^{2},
\end{align}where  $(c)$ follows from \eqref{eqn: admm-update2_c},  $(b)$ holds because $ \rho \geq \sqrt{\gamma_i^{t}}$, which implies
\begin{align}
\frac{1}{\beta^t}=& \frac{\gamma_i^t}{(\rho + \gamma_i^{t-1})^{2}}
\leq \frac{\gamma_i^t}{\rho^{2} }
\leq  1.
\end{align}
$\hfill\blacksquare$

\end{onecolumn}

\end{document}